\documentclass[a4paper,fleqn]{cas-dc}

\usepackage[numbers]{natbib}
\def\tsc#1{\csdef{#1}{\textsc{\lowercase{#1}}\xspace}}
\tsc{WGM}
\tsc{QE}
\tsc{EP}
\tsc{PMS}
\tsc{BEC}
\tsc{DE}

\usepackage{hhline}
\usepackage{algorithmic}
\usepackage{algorithm}  
\graphicspath{{Figs/}}
\usepackage{multirow}

\usepackage{multicol}

\usepackage{makecell}
\newcommand{\guan}[1]{\textcolor{blue}{#1}}

\begin{document}
\let\WriteBookmarks\relax
\def\floatpagepagefraction{1}
\def\textpagefraction{.001}

\shorttitle{Multimodal Zero-Shot Learning for Tactile Texture Recognition}

\shortauthors{G. Cao et~al.}

\title [mode = title]{Multimodal Zero-Shot Learning for Tactile Texture Recognition}                      
% \tnotemark[1,2]

\tnotetext[1]{This work was funded in part by the EPSRC project ``ViTac: Visual-Tactile Synergy for Handling Flexible Materials'' (EP/T033517/2).}

\author[1]{Guanqun Cao}[orcid = 0000-0002-3462-2567]

\affiliation[1]{{Department of Computer Science, University of Liverpool,}, 
    city={Liverpool},
    % citysep={}, % Uncomment if no comma needed between city and postcode
    postcode={L69 3BX}, 
    country={United Kingdom}}

\author[2]{Jiaqi Jiang}

\affiliation[2]{{Department of Engineering, King's College London,}, 
    city={London},
    % citysep={}, % Uncomment if no comma needed between city and postcode
    postcode={WC2R 2LS}, 
    country={United Kingdom}}

\author[1]{Danushka Bollegala}

\author[3]{Min Li}
\affiliation[3]{{School of Mechanical Engineering, Xi'an Jiaotong University,}, 
    city={Xi'an},
    % citysep={}, % Uncomment if no comma needed between city and postcode
    postcode={710049}, 
    country={China}}

\author[2]{Shan Luo}
\ead{shan.luo@kcl.ac.uk}
\cormark[1]

\cortext[cor1]{Corresponding author}
% \cortext[cor2]{Principal corresponding author}

% \fntext[fn1]{This is the first author footnote. but is common to third
% \fntext[fn2]{Another author footnote, this is a very long footnote and
%   it should be a really long footnote. But this footnote is not yet
%   sufficiently long enough to make two lines of footnote text.}

% \nonumnote{This note has no numbers. In this work we demonstrate $a_b$
%   the formation Y\_1 of a new type of polariton on the interface
%   between a cuprous oxide slab and a polystyrene micro-sphere placed

\begin{abstract}
%\shan{Follow this framework to write the Abstract: \\
%1. Background \\
\guan{Tactile sensing plays an irreplaceable role in robotic material recognition. It enables robots to distinguish material properties such as their local geometry and textures, especially for materials like textiles.
%2. Unmet needs \\
However, most tactile recognition methods can only classify known materials that have been touched and trained with tactile data, yet cannot classify unknown materials that are not trained with tactile data. 
%3. Proposed solution \\
% To solve this problem, we propose a multimodal zero-shot learning framework to recognise materials that have not been touched by the robot before. 
To solve this problem, we propose a tactile zero-shot learning framework to recognise unknown materials when they are touched for the first time without requiring training tactile samples.
% based on generative models.
%4. Rationale \\
%5. Experiment results \\
%6. Impact or Importance of your work scientifically \\
%Currently, we only have 1 and 2.
%}
% Using tactile sensing in the recognition of materials by robots plays an important role in the textile recycling industry. 
% For robots to recognise materials, tactile sensing can provide information regarding local geometry, micro-structure, and textures to understand the properties of objects, which plays an important role in the material recognition.
% % However, current recognition methods suffer from it can only recognise the materials included in the training set , i.e., it is difficult for the robots to recognise the materials never touched before. 
% However, current methods of tactile recognition are limited to recognizing only materials that have already been trained, i.e., it is difficult to recognise materials the robot has never handled before.
% As new materials are being developed continuously, it is impossible to include data from all materials into training set.
The visual modality, providing tactile cues from sight, and semantic attributes, giving high-level characteristics, are combined together to bridge the gap between touched classes and untouched classes.
A generative model is learnt to synthesise tactile features according to corresponding visual images and semantic embeddings, and then a classifier can be trained using the synthesised tactile features of untouched materials for zero-shot recognition.
Extensive experiments demonstrate that our proposed multimodal generative model can achieve a high recognition accuracy of $83.06\%$ in classifying materials that were not touched before. 
The robotic experiment demo and the dataset are available at \textit{https://sites.google.com/view/multimodalzsl}}
\end{abstract}

% \includegraphics{figs/grabs.pdf}

% % Research highlights
% \begin{highlights}
% \item Research highlights item 1
% \item Research highlights item 2
% \item Research highlights item 3
% \end{highlights}

% Keywords

\maketitle

\section{Introduction}

The material properties of the object's surface, such as roughness, texture, and hardness, are key information for robots to interact with their surroundings.
As such, recognising the surface materials, which allows robots to be aware of the object categories and properties, is fundamental to many manipulation tasks, such as grasping~\citep{chitta2011tactile}, robot navigation~\citep{roy1996surface}, and social and collaborative interactions~\citep{hassan2022tactile}.

The most widely used methods for material recognition are based on vision as it provides shapes, colours, and appearances to perceive properties of different materials~\citep{liu2010exploring}.
Vision-based methods, however, are subject to lighting and occlusion~\citep{lecun2004learning,lowe1999object}. 
Moreover, due to the enormous range of appearances that a single material might exhibit, it is difficult to establish distinguishable representations from vision.  
% To address this problem, semantic attributes, for example, ``soft'', ``lightweight'' and ``smooth'' to describe a nightshirt, have been introduced for \shan{a unified representation} for material recognition~\citep{cimpoi2014describing,zhai2019deep}.
To address this problem, the semantic attributes (e.g., ``knitted'' and ``fibrous'' to describe the wool) have been introduced as complimentary information to assist visual material recognition~\citep{cimpoi2014describing,zhai2019deep}.
% Nevertheless, \shan{semantic attributes of the same material may be different due to different fabrication methods, biasing the results of material recognition.}
%Nevertheless, \guan{semantic attributes are limited to the number of attributes used for representation and human bias during the annotation.}

% Unlike vision and semantic attributes,
% tactile sensing can measure the micro-structures of the object's surface that \shan{cannot be easily altered} even if the appearance and shape are changed, which allows the robot to recognise different materials effectively~\citep{luo2017robotic}.
Unlike vision and semantic attributes,
tactile sensing can measure the micro-structures of the object's surface even if the appearance and shape are changed, which allows the robot to recognise different materials effectively~\citep{luo2017robotic}.
Furthermore, many exclusive physical properties that cannot be obtained in other sensory modalities, such as friction and compressibility, can be measured by tactile sensing through rich physical interaction, giving a good understanding of different materials.

In recent studies on material recognition by tactile sensing, a large amount of tactile data need to be collected first, and then a projection function between the collected tactile data and material classes is learnt with optimisation algorithms or machine learning methods~\citep{giguere2011simple,kerr2018material}. By using the learnt projection function, robots can predict the classes of the contacted material. However, there are two main issues limiting the application of such methods: 1) The material to be recognised must be known and included in the classes of the training dataset, which is hard to be met due to the continuous development of new materials. 2) The collection of a large amount of tactile data is costly as the delicate tactile sensors are easily damaged after numerous physical contacts and data collection could be time-consuming %due to \guan{a large number of explorations}.
% In most of previous material recognition methods with tactile sensing, robots need first to touch the target materials with a tactile sensor to collect a large amount of data, then a projection function will be learnt between the collected tactile data and material classes, using optimization algorithms or machine learning methods~\citep{giguere2011simple,kerr2018material}. 
% Finally, robots are able to predict the classes of test tactile data with the projection function. 
% These methods have a limitation that the classes of the test set must be included in the training set.
% % Otherwise, the robot will probably misclassify the new data as a exiting touched class.
% Since new materials are being developed continuously, it is impossible to include data from all kinds of materials into the training data.
% On the other hand, the collection of tactile data is also a difficult task.
% Tactile sensors are likely to be damaged because of frequent physical contacts for data collection.
% % Moreover, due to the difficult exploration process and dynamic environment, tactile data collection is very time consuming and expensive.
% Furthermore, the complexity of exploration, such as the dynamic environment, makes tactile data collection time-consuming and costly.
% As a result, the insufficient tactile samples for training and the lack of annotation will present the tactile recognition with great challenges.

As a result, the lack of tactile samples for training and the absence of annotations pose challenges to recognise the materials never touched by robots before.
Hence, Zero-Shot Learning (ZSL) for tactile recognition, i.e., to build a model using the tactile data of known materials (i.e., ones touched by the robot before) to recognise unknown (untouched) materials that lack training tactile samples, is desired. 
% \shan{a sentence should be added here this capability is enabled with other sensing modalities.}
This capability can be acquired by obtaining a shared subspace (e.g., with the visual domain) to transfer the knowledge learnt from touched materials to untouched materials.

% To solve this problem, we propose a tactile zero-shot learning framework to recognise unknown materials when they are touched for the first time without requiring tactile training samples.
% Even though existing methods such as sim2real applied by a simulation model can provide a large amount of data to train with~\citep{gomes2021generation,agarwal2021simulation}, there is still a large gap between simulated tactile data and real tactile data.

A good example of tactile ZSL is the recognition of daily fabrics, which provide a variety of appearances, physical properties, and tactile feelings.
For humans, it is an easy task to recognise a new material by the sense of touch based on our prior knowledge and descriptions.
For example, if we are given a description of silk ``\textit{the silk material is usually very smooth, soft and cool}'', we can recognise a fabric made of silk even when it is the first time for us to touch a piece of silk.
This capability to recognise new materials that were never touched before is due to that our brain is able to transfer the knowledge gained in one sensory modality to another, i.e., cross-modal transfer~\citep{cole1961cross}.
Similarly, humans can also \textit{imagine} a tactile feeling when \textit{observing} the materials~\citep{bower1970coordination}, i.e., \textit{seeing to feel}~\citep{lee2019touching}.

\begin{figure}[t]
	\centering
	\includegraphics[width=1\columnwidth]{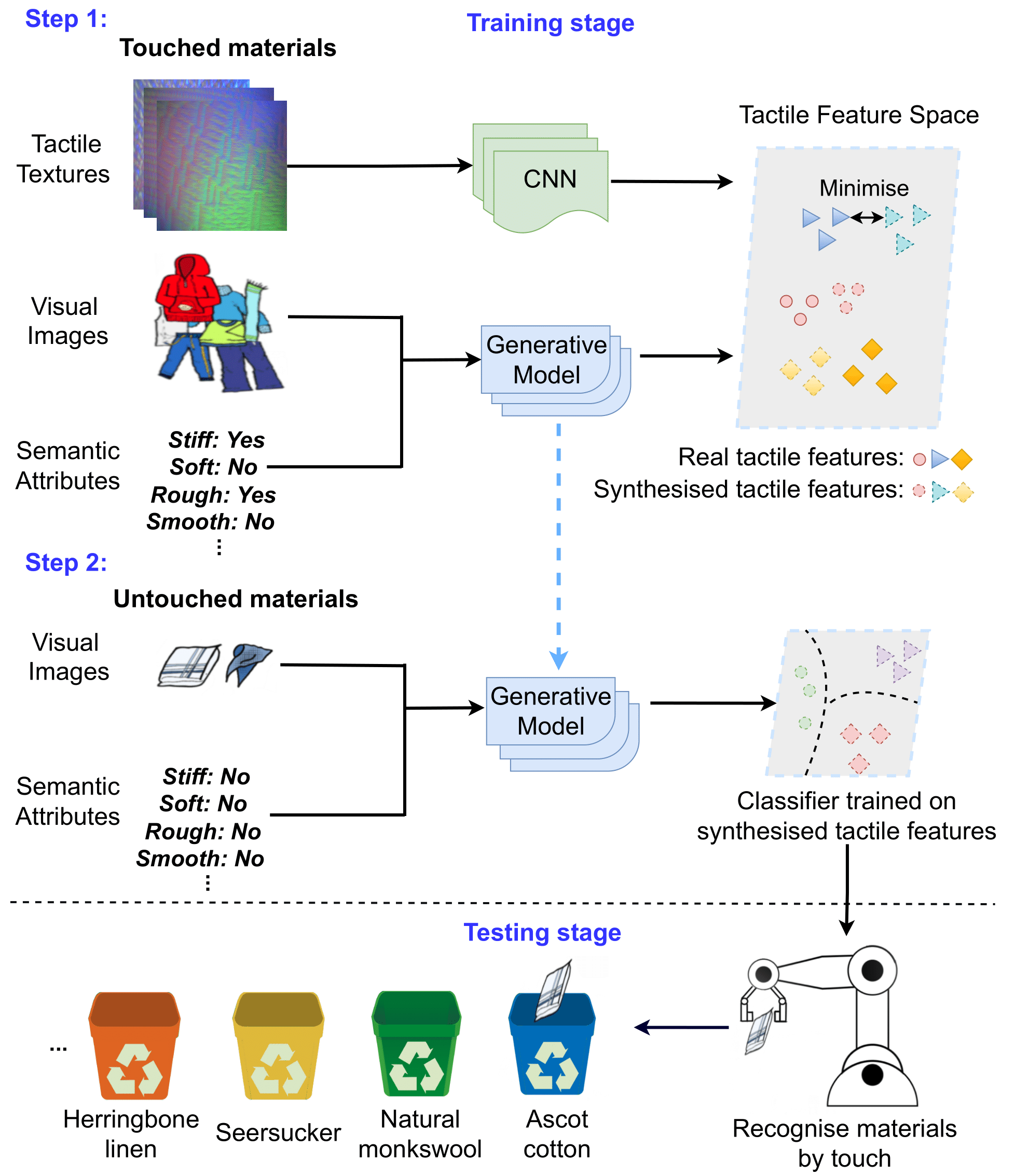}  
    \caption{{\textbf{Tactile Zero-Shot Learning for material recognition.}  Training stage: Step 1: Given visual images, semantic attributes and tactile textures of known (touched) materials, a generative model is trained to synthesise tactile features by minimising the distance between the distributions of real tactile features and synthesised tactile features. Step 2: A classifier is trained by the synthesised tactile features of unknown (untouched) materials.
    Testing stage: By using the classifier, the robot is able to recognise and sort unknown materials by tactile sensing. }}
	\label{fig:industry}
	% \vspace{-2.0em}
\end{figure}

% To achieve this goal, the learnt knowledge is transferred from touched classes to untouched classes.
Inspired by the above, we propose a multimodal Zero-Shot Learning approach for tactile textures recognition that employs both visual information and semantic attributes from fabrics to synthesise corresponding tactile features.
% To bridge the gap between touched and untouched materials, auxiliary visual information and semantic attributes are implemented as shared subspaces.
As shown in Fig.~\ref{fig:industry}, firstly, a generative model is trained to synthesise tactile features with touched materials. 
After training, the generative model is used to synthesise tactile features of untouched materials using corresponding visual information and semantic attributes.
Then, a classifier is trained on the synthesised tactile features of untouched materials.
Finally, the robot can recognise these untouched materials by tactile sensing. 
Our proposed tactile ZSL method addresses a practical problem in material recognition using robots: even if tactile data is not available to train the robot for new materials, they can still be recognised and sorted when being touched for the first time.

The contributions of this paper can be summarised as follows:
\begin{enumerate}

    \item We propose a multimodal zero-shot learning framework to recognise materials that have not been touched before;
    % \item  Visual and semantic modalities are combined together to synthesise tactile features to achieve a synergistic effect in tactile Zero-Shot Learning, for the first time;
    \item  We propose to synthesise tactile features from visual images and semantic attributes to achieve a synergistic effect in tactile zero-shot learning, which is the first of its kind;
    \item We collect a new dataset from 50 pieces of fabrics to train the model, including visual images, semantic attributes and tactile data, and validate our proposed method on the untouched materials.
\end{enumerate}

The remainder of this paper is structured as follows: Section \ref{relatedwork} reviews the related works; Section \ref{formulation} introduces the problem formulation of tactile Zero-Shot Learning;  Section \ref{methods} details the multimodal framework for Zero-Shot Learning of tactile recognition; Section \ref{datacollection} introduces the experimental setup; Section \ref{experiments} shows the experimental results; 
Section \ref{discussion} discusses several aspects of tactile Zero-Shot Learning; 
Finally, Section \ref{conclusions} summarises the paper and gives conclusions.

\section{Related Works}\label{relatedwork}
In this section, we will first review works on material recognition with tactile textures, followed by discussions of Zero-Shot Learning for visual recognition and tactile recognition respectively.

\subsection{Material Recognition with Tactile Textures}

Tactile textures are crucial in understanding the properties of materials since they convey important information of local geometry and micro-structures of the contacting object's surface.
% A surface's texture can be determined by measuring friction coefficients, roughness and micro-structure patterns.
% In the first two cases, both of them can be approached a force sensor or tactile sensor. 
With the development of tactile sensing, various tactile sensors based on different sensing technologies, such as microphones~\citep{luo2017knock,edwards2008extracting}, strain gauges~\citep{jamali2011majority}, MEMS~\citep{de2009tactile,kim2005texture}, capacitive~\citep{taunyazov2019towards,roberge2021tactile}, and piezoresistive~\citep{guo2021visual}, have been implemented in the tactile texture recognition.
Recently, thanks to their high resolution and low cost, camera-based optical tactile sensors, such as the GelSight sensor~\citep{yuan2017gelsight} and the TacTip sensor~\citep{ward2018tactip}, have been implemented in the material perception.
% In~\citep{li2013sensing}, the tactile textures are classified based on the local binary patterns with the tactile images collected from GelSight sensor.
In~\citep{yuan2018active}, the GelSight is applied in an autonomous tactile exploration that enables the robots to perceive material properties.
In~\citep{cao2020spatio}, a spatio-temporal attention mechanism is proposed to emphasise the salient features to recognise textures present in the tactile image sequences collected by a GelSight sensor.
% In \citep{luo2018vitac}, a joint latent space is used for sharing features between vision and touch to improve the performance of cloth material recognition, also with the GelSight sensor.
In \citep{luo2018vitac}, a joint latent space of vision and touch sensing for sharing features is learnt to improve the cloth material recognition.
However, these methods are limited to classifying known materials and cannot recognise unknown materials that are not included in the training classes.

\subsection{Zero-Shot Learning for Visual Recognition}
An increasing interest has been shown in recognising unseen objects based on visual images without any training examples, i.e., Zero-Shot Learning.
The use of semantic embeddings learnt from semantic attributes is common to bridge the gap between the seen classes and the unseen classes.
A projection function can be learnt, for instance, from visual to semantic space~\citep{lampert2013attribute,socher2013zero}, from semantic to visual space~\citep{das2019zero,zhang2017learning}, or a shared latent space~\citep{akata2013label,lei2015predicting}, to connect vision and semantic information for the recognition.
However, as the classes of the seen data and unseen data can be unrelated, the data distributions may be different.
If the projection function that is learnt from seen data is applied to the unseen data directly, it may generate unknown bias, known as the domain shift problem~\citep{fu2015transductive}.
% the projection methods may encounter a domain shift problem~\citep{fu2015transductive}.
Another popular method is based on the generative model~\citep{xian2018feature,felix2018multi}, where the visual features of unseen classes are synthesised using semantic information, and synthesised visual features are used to train a classifier to recognise unseen data, which alleviates the domain shift problem significantly. 
%Subsequently, the recognition will become a supervised learning problem that a classifier will be trained based on the synthesised feature to classify the novel data, which is able to mitigate the problem of domain shift.
However, these studies focus on the visual ZSL problem and there have been no works on tactile ZSL based on generative models that use visual images and semantic attributes together to synthesise tactile features.

\subsection{Zero-Shot Learning for Tactile Recognition}

Due to the difficulty of the tactile data collection and annotation, a great demand exists for tactile ZSL. However, compared with the visual ZSL, the ZSL problem for tactile recognition has been much less investigated.
In~\citep{liu2018cross}, visual images are used as the auxiliary information to connect the touched objects and untouched objects with dictionary learning.
% In~\citep{abderrahmane2018haptic}, the Direct Attributes Prediction (DAP) is applied in the tactile ZSL, by using tactile data to predict semantic attributes.
In~\citep{abderrahmane2018haptic}, the semantic attributes are predicted using tactile data with Direct Attributes Prediction (DAP), and the corresponding categories of untouched materials can be determined by the predicted attributes.
In~\citep{abderrahmane2018visuo}, semantic attributes are learnt from both visual data and tactile data using DAP. 
In~\citep{abderrahmane2019deep}, a generative model is developed to synthesise tactile features with semantic attribute inputs for tactile ZSL.
% However, \shan{there has been no work using the generative model conditioned on visual images in tactile ZSL.}
Although many studies have demonstrated the superiority of the generative model-based method in ZSL~\citep{ye2021alleviating, li2020learning} and the visual modality is able to provide an objective measurement for the target object, there has been no work using the generative model-based method conditioned on visual images in tactile ZSL.
Moreover, to the best of the authors’ knowledge, the multimodal generative framework, conditioned on both visual images and semantic attributes, has not been attempted before for tactile ZSL.
In this work, we propose a multimodal generative model that integrates visual images and semantic attributes to synthesise tactile features for zero-shot tactile recognition, for the first time.

\section{Problem Formulation}\label{formulation}

% \begin{figure}
% 	\centering
% 	\includegraphics[width=1\columnwidth]{casestudy_5.pdf}
% % 	\caption{\textbf{\textit{Tactile selective attention.}} If pressing against the Lego bricks to distinguish what it is, the studs on the green Lego brick will provide more cues than the flat region on the red one.   }
%     % \captionsetup{font={small,bf,stretch=1.25},justification=raggedright}
%     \captionsetup{font={small,stretch=1.25}}
% 	\caption{\textbf{The process of multimodal tactile ZSL.} In tactile ZSL, tactile data of untouched classes is recognised by using auxiliary information, i.e., visual images and semantic attributes based on the experience from the touched classes.}
% 	\label{fig:first}
% % 	\vspace{-2.0em}
% \end{figure}

\begin{figure}[t]
	\centering
	\includegraphics[width=1\columnwidth]{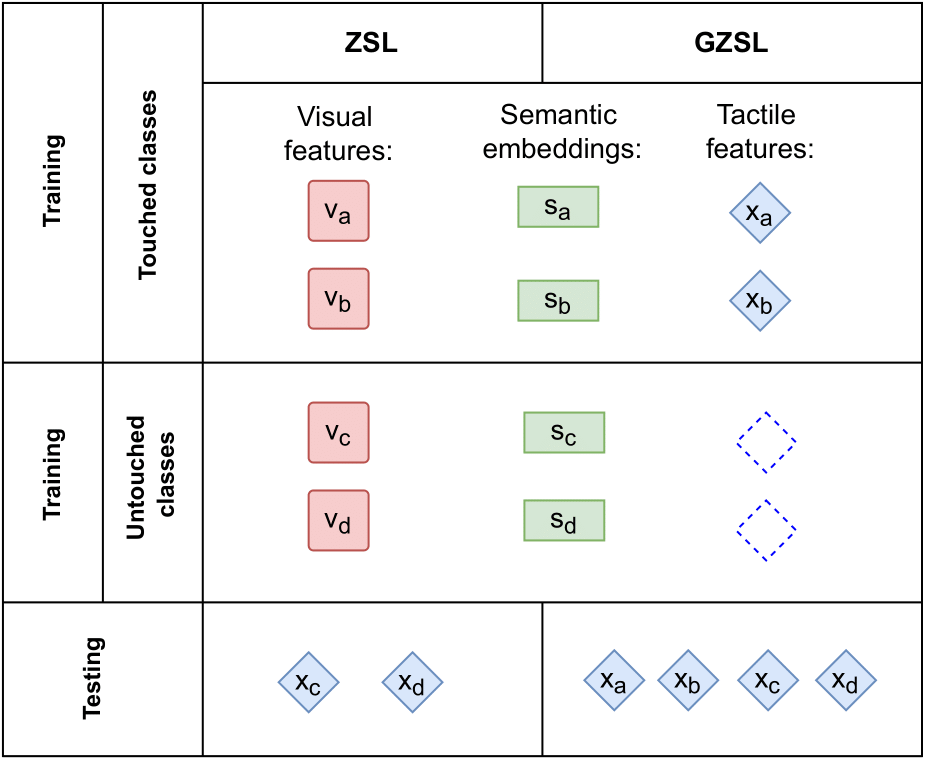}
    % \captionsetup{font={small,stretch=1.25}}
	\caption{ \textbf{The different configurations between conventional ZSL and GZSL.} In the training, visual features, semantic embeddings and tactile features are available for touched classes, whereas only visual features and semantic embeddings are available for untouched classes; in the testing stage, in conventional ZSL, only the tactile features from untouched classes will be tested. In GZSL, the tactile features from both touched classes and untouched classes will be tested.  }
	\label{fig:zslgzsl}
	% \vspace{-1em}
\end{figure}

% In our tactile ZSL, the data includes the touched (training) set and untouched (testing) set.
% The touched set can be represented by $D_{t}=\left\{\left(x_{t}, v_{t}, s_{t}, y_{t}\right) \mid x_{t} \in X_{t}, v_{t} \in V_{t}, s_{t} \in S_{t}, y_{t} \in Y_{t}\right\}$ where $x_{t} \in \mathbb{R}^{d}$ stands for the $d$-dimensional tactile features from tactile feature space $X$, $v_{t} \in \mathbb{R}^{d}$ represents the $d$-dimensional visual features from visual feature space $V$, $s_{t} \in \mathbb{R}^{k}$ denotes k-dimensional the semantic features from semantic feature space $S$, and  
% The untouched set is given as $D_{u}=\left\{\left(x_{u}, v_{u}, s_{u}, y_{u}\right) \mid x_{u} \in X_{u}, v_{u} \in V_{u}, s_{u} \in S_{u}, y_{u} \in Y_{u}\right\}$, where $x_{u}$, $v_{u}$, $s_{u}$ denote the untouched tactile features, visual features, semantic features respectively. 
% $Y_{u}=\left\{y_{u}^{1}, y_{u}^{2}, \ldots, y_{u}^{n_{u}}\right\}$ indicates the label set of untouched classes.
The tactile ZSL aims to use tactile sensing to recognise new materials that have no tactile samples during the training process, when they are touched for the first time.
Specifically, tactile ZSL can be divided into two stages: the training stage and the inference (testing) stage.
The materials to be recognised can also be split into two sets: the touched classes that are touched by the robot during training and the untouched classes that are not touched in training and will be used in the test.
To satisfy the zero-shot assumption that there are no tactile training samples from untouched classes, the tactile ZSL requires training a model only on the touched classes, and classifying the tactile data of untouched classes.
% An essential component of the tactile ZSL is to bridge the gap between touched and untouched classes with \shan{auxiliary information.}

An essential component of the tactile ZSL is to use auxiliary information, which can provide additional characteristics of the target object (such as semantic attributes), to bridge the gap between touched and untouched classes.
In our multimodal ZSL framework, we use visual images and semantic attributes from both touched materials and untouched materials as auxiliary information and recognise the tactile data of untouched classes based on the knowledge learnt from the touched classes.

Let $x \in \mathbb{R}^{d_x}$, $v \in \mathbb{R}^{d_v}$, $s \in \mathbb{R}^{d_s}$ represent the tactile features, the visual features, and the semantic embeddings learnt from tactile data $X$, visual images $V$ and semantic attributes $S$ respectively.
$Y=\left\{y^{1}, y^{2}, \ldots, y^{n}\right\}$ denotes the corresponding label set. 
% In tactile ZSL, the data includes the touched set and untouched set.
We use subscripts $t$ and $u$ to indicate the touched materials and the untouched materials respectively.
The data of touched set can be denoted as $D_{t}=\left\{(x_{t}, v_{t}, s_{t}, y_{t})\right\}$. %\right) \mid x_{t} \in X_{t}, v_{t}$. %\in V_{t}, s_{t} \in S_{t}, y_{t} \in Y_{t}\right\}$. 
The data of untouched set can be denoted as $D_{u}=\left\{\left(x_{u}, v_{u}, s_{u}, y_{u}\right)\right\}$. % \mid x_{u} \in X_{u}, v_{u} \in V_{u}, s_{u} \in S_{u}, y_{u} \in Y_{u}\right\}$.
In ZSL, touched classes and untouched classes are disjoint, which means that $Y_{t} \cap Y_{u}=\emptyset$.

Following~\citep{chao2016empirical}, we formalise our tactile ZSL in two different settings: one in conventional ZSL and the other in Generalised Zero-Shot Learning (GZSL). 
As an example is shown in Fig.~\ref{fig:zslgzsl}, there are two touched materials and two untouched materials, and we use the subscripts \textit{a, b, c, d} to represent them respectively. 
In the training stage of conventional ZSL, visual features, semantic embeddings and tactile features are available for materials \textit{a} and \textit{b}, while only visual features and semantic embeddings are available for materials \textit{c} and \textit{d}. 
The tactile features of materials \textit{c} and \textit{d} are absent in the training stage and is only used in the testing stage. 
In GZSL, the data availability is the same with the ZSL setting in the training stage. 
The main difference between the conventional ZSL and GZSL comes from the testing stage: in GZSL, the tactile features from all materials \textit{a}, \textit{b}, \textit{c} and  \textit{d} will be tested.

To summarise, given $D_t$ and $D_u$, in conventional ZSL, a classifier $f_{Z S L}: X_u \rightarrow Y_{u}$ will be learnt to recognise the tactile data of untouched materials, and GZSL requires learning a classifier $f_{G Z S L}: X \rightarrow Y_{t} \cup Y_{u}$ to classify the tactile data from both touched and untouched materials.
Note that in both cases $X_{u}$ is not available during training, and is only used in the test.
The reasons why we have two settings of ZSL are that: if given the prior knowledge that which material is unknown, we only need to recognise the unknown materials that never been touched before, which is a ZSL setting; in practical applications, unknown materials may be mixed with previously known materials, which is a typical GZSL setting.

%As shown in Fig.~\ref{fig:zslgzsl}, a ZSL problem can be categorized into the conventional ZSL and generalized zero-shot learning (GZSL).
%Given $D_t$ and $D_u$, in conventional ZSL, a classifier $f_{Z S L}: X_u \rightarrow Y_{u}$ will be learnt to recognise the tactile data of untouched classes, and GZSL requires learning a classifier $f_{G Z S L}: X \rightarrow Y_{t} \cup Y_{u}$ to classify the tactile data from both touched and untouched classes.
%Note that in both cases $X_{u}$ is not available during training, and is only used in the test. 
%We will test our method on both ZSL and GZSL settings.

% Specifically, ZSL can be divided into inductive method and transductive method.
% Inductive methods only use labeled data from training set for training, whereas transductive methods use both labeled data from training set and unlabeled data from testing set during training.
% ZSL can also be categorized into the conventional ZSL and generalized zero-shot learning (GZSL).
% In conventional ZSL, a classifier $f_{Z S L}: X \rightarrow Y_{u}$ will be learnt to recognise the untouched tactile features, and the GZSL requires learning a classifier $f_{G Z S L}: X \rightarrow Y_{t} \cup Y_{u}$ to classify the both touched tactile features and untouched tactile features.

% Our proposed generative method is trained within the inductive setting, and the method will be tested both on ZSL and GZSL settings.

\begin{figure*}[t]
	\centering
	\includegraphics[width=2\columnwidth]{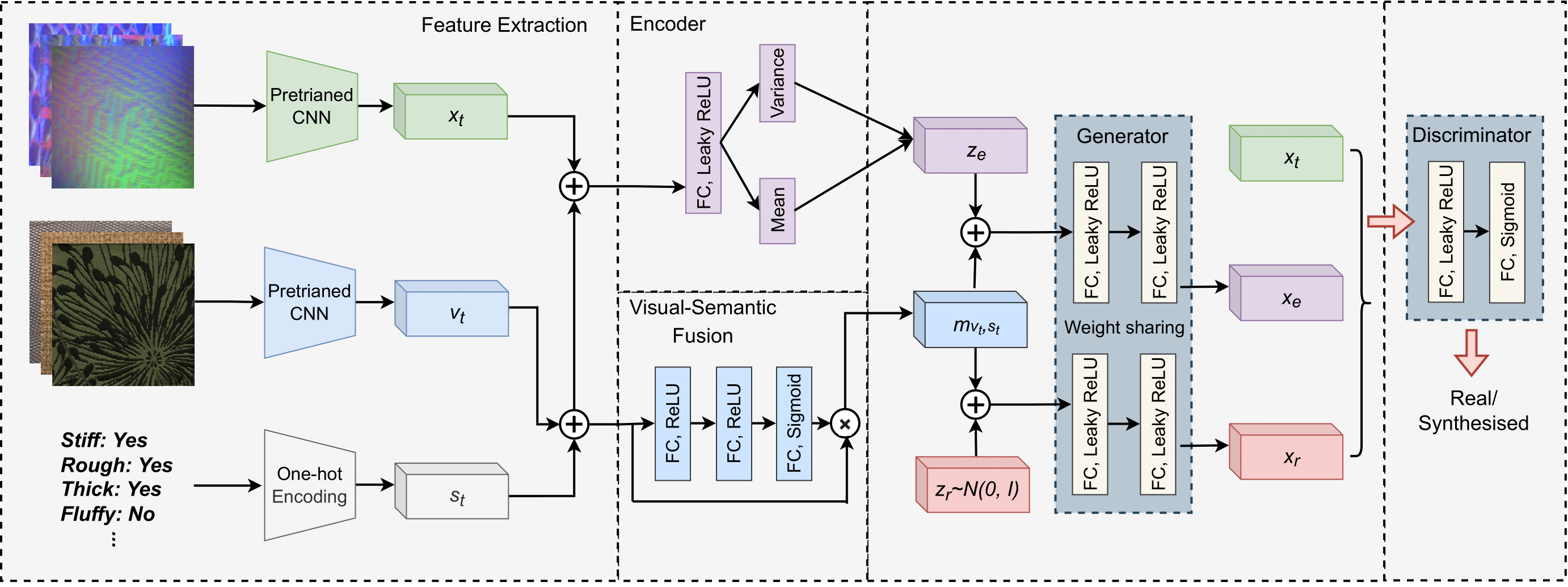}
% 	\captionsetup{font={small,stretch=1.25}}
	\caption{\textbf{Illustration of the proposed VS2T-ZSL generative model.} Our generative model consists of five modules: (1) a feature extraction module; (2) an encoder module $E$; (3) a visual-semantic fusion module VS-F; (3) a generator module $G$; (4) a discriminator module $D$. 
% 	Our generative model is trained using the touched tactile textures with corresponding visual images and semantic attributes.
% In our model, $E$ encodes the input features to a continuous latent space; VS-F highlights salient features across the visual domain and the semantic domain; $G$ is used to reconstruct tactile features; $D$ discriminates synthesised tactile features from real features. 
	After training, the generator $G$ is used to synthesise the tactile features of untouched materials based on corresponding visual images and semantic attributes.     
	} 
	\label{fig:framework}
	\vspace{-1.2em}
\end{figure*}

\section{Methodologies}\label{methods}

Our proposed multimodal tactile ZSL framework that learns tactile features from visual images and semantic attributes, i.e., VS2T-ZSL, consists of two main components: a generative model to synthesise tactile features of untouched materials using auxiliary information, and a recognition model that is trained with synthesised features to recognise the real tactile textures from untouched materials.

As illustrated in Fig.~\ref{fig:framework}, our proposed generative model is a combination of  Variational Autoencoder (VAE) and Generative Adversarial Network (GAN)~\citep{bao2017cvae}. VAE consists of an encoder and a decoder where an explicit distribution can be obtained by encoding the data to a multi-dimensional Gaussian distribution.
However, VAE often generates blurry tactile results due to the limitations of reconstruction loss in the decoder ~\citep{cai2019multi}. GAN usually contains a generator and a discriminator. In contrast to VAE, GAN learns an implicit distribution where the discriminator is applied to evaluate the quality of the synthesised tactile results from the generator to get sharp and clear features~\citep{goodfellow2014generative}.
However, modal collapse~\citep{tran2018dist} may occur during the training of GAN where the generator of network produces only specific outputs regardless of the inputs. To this end, we combine VAE and GAN in our tactile ZSL framework to perform an efficient training and clear generated results.

%Each component of the generative model will be detailed in Subsection ~\ref{generative}.

% The generative model is trained by visual features, tactile features, and semantic embeddings from touched set.
After training our joint generative model, the generator $G$ is used to synthesise the tactile features of untouched materials from corresponding visual images and semantic attributes.
% To recognise the tactile features from untouched materials, a Softmax classifier will be trained based on their synthesised tactile features to recognise real tactile features.
In conventional ZSL, a classifier $CLS_{u}$ will be trained on the synthesised untouched tactile features to recognise the real untouched tactile features.  
% In the GZSL setting, the challenge lies in that we do not know if the test data is from the touched set or the untouched set. 
% Similar to the methods in~\citep{abderrahmane2019deep,socher2013zero}, we apply a Gaussian distribution to measure the probability density distribution of the touched data for novelty classification. 
% An input sample is categorized as touched if it locates in a high-density region, and as untouched if it does not.
In GZSL setting, apart from $CLS_{u}$, we train another classifier $CLS_{t}$ with the touched set to recognise the touched tactile data.
More details will be discussed in Subsection~\ref{recognition}.
% Given the input test data from touched or untouched set, the test data can be fed into corresponding classifier for recognition.

\subsection{Tactile Feature Generative Model}\label{generative}

% In this section, we will discuss each component's loss function, including the encoder loss, reconstruction loss, discriminator loss, classifier loss, and generator loss. 
In this section, our tactile feature generative model will be introduced. As illustrated in Fig.~\ref{fig:framework}, our generative model consists of 1) a feature extraction module that extracts the high dimensional features from visual images, tactile textures, and semantic attributes; 2) an encoder $E$ that maps the tactile features and auxiliary information to a continuous latent space; 3) a visual-semantic fusion module $VS-F$ that emphasises the salient features across the visual domain and the semantic domain for generation process; 4) a generator (decoder) $G$ that samples from the latent space to reconstruct tactile features conditioned on fused features; 5) and a discriminator $D$ that discriminates if the input is a synthesised tactile feature or a real tactile feature (which is extracted from the tactile texture).

%it consists of a feature extraction module, an encoder ($E$) module, a visual-semantic fusion ($VS-F$) module, a generator ($G$) module and a discriminator ($D$) module. 

\noindent \textbf{Feature extraction module.} Instead of synthesising the tactile textures directly, we focus on generating high dimensional tactile features for the tactile recognition task. 
Firstly, we extract the features from visual images $V_t$, tactile textures $X_t$, and semantic attributes $S_t$ to represent the material from different domains.
Specifically, two pretrained ResNet50 models~\citep{he2016deep} are fine-tuned using the visual images and the tactile textures from touched materials respectively, and visual features $v_t$ and tactile features $x_t$ are extracted from the last pooling layer of the fine-tuned models.
To obtain semantic representations, we use the one-hot encoding~\citep{hackeling2017mastering} with semantic attributes to get semantic embeddings $s_t$. 

\noindent \textbf{Encoder module.} Similar to VAE, the encoder $E$ encodes input features and outputs the mean vector and the variance vector of latent space. We minimise the Kullback–Leibler (KL) divergence between the output latent distribution $q\left(z \mid x_{t}, v_{t}, s_{t}; \phi_{E}  \right)$ and the standard normal distribution $p(z)$ to ensure a continuous latent space for the generative process:
\begin{equation}
\mathcal{L}_{K L}=K L\left(q\left(z \mid x_{t}, v_{t}, s_{t}; \phi_{E}  \right) \| p\left(z\right)\right),
\label{eq:1}
\end{equation}
where %$x,v,s$ represent paired tactile features, visual features and semantic features extracted from the tactile textures, visual images, and semantic attributes respectively, and 
$\phi_{E}$ is the parameters of the encoder, and $K L\left( \cdot \right)$ represents the KL divergence.
%The minimization of the KL loss between the output distribution and normal distribution ensures a regularised latent space for generative process.
Compared with a single GAN, the encoder in our joint model makes a continuous space for generation, which enables a better generalisation ability to the untouched materials while synthesising their tactile features.

\noindent \textbf{Visual-semantic fusion module.} 
% The generator is used to synthesise the tactile features with visual features and semantic attributes. 
As the generator has multimodal input of heterogeneous sources for the generation of tactile features, a simple concatenation of visual features and semantic features is not sufficient in practice. 
Inspired by~\citep{hu2018squeeze}, we design a visual-semantic fusion function to emphasise the task-relevant features across different modalities. 
The fused features can be given by:
\begin{equation}
\begin{aligned}
m_{v_{t},s_{t}} &= f\otimes(\sigma\left({W}_{3}\delta\left({W}_{2}\delta\left({W}_{1}f\right)\right)\right)),
\end{aligned}
\end{equation}
where $f = \text{Concat}(v_{t},s_{t})$, ${W}_{1},{W}_{2},{W}_{3}$ are learnable matrices, which are implemented by fully connected (FC) layers. 
$\delta$ and $\sigma$ represent a ReLU and a Sigmoid activation functions respectively, and $\otimes$ denotes element-wise product. 
Compared with a simple concatenation, our visual-semantic fusion module is able to highlight salient features and suppress redundant features in visual and semantic modalities, by assigning different weights to the feature vector.

\noindent \textbf{Generator module.} The generator $G$ tries to reconstruct the tactile features using latent vectors with fused features.
For the generator $G$, we minimise $\ell_{2}$ reconstruction loss and pairwise feature matching loss~\citep{bao2017cvae} for feature reconstruction:
 
\begin{equation}
\begin{aligned}
\mathcal{L}_{rec}&=\mathbb{E}_{x_t,x_e}\left\|{x}_{t}-{x}_{e}\right\|_{2}^{2}+\mathbb{E}_{x_t,x_e}\left\|f_{D}(x_t)-f_{D}({x}_{e})\right\|_{2}^{2},
\end{aligned}
\label{eq:2}
\end{equation}
where $x_{e} = G(z_{e}, m_{v_t,s_t})\in \mathbb{R}^{d_x}$ represents the synthesised tactile features, and $z_{e}\sim q\left(z \mid x_t, v_t, s_t; \phi_{E}  \right) $ denotes the latent vectors sampled from the latent distribution. $f_{D}$ denotes the outputs of the last hidden layer from the discriminator.
% This function enables the synthesised results $x^{\prime}$ close to the input $x$.

% However, a disadvantage of VAE is that the generated results are not clear due to the imperfect reconstruction function.
% To solve this problem, we further apply a discriminator to improve the quality of generated features.
% In the traditional GAN network, the generator is applied to generate the realistic data that close to the real data, and the discriminator is used to determine if the input is from real domain or fake domain.
% The discriminator is used to discriminate the real/fake features.
% In our task, the discriminator $D$ aims to minimize the following loss:
% Given synthesised features and real tactile features, discriminator $D$ tries to minimize the loss:

\noindent \textbf{Discriminator module.} The discriminator $D$ is used to identify synthesised tactile features from real tactile features. 
Concretely, the discriminator is learnt by minimising the loss:
\begin{equation}\label{eq:3}
\begin{aligned}
\mathcal{L}_{D}=&-\mathbb{E}_{x_t}[\log D(x_t)]-\mathbb{E}_{x_e}[\log (1-D(x_{e}))] \\
&-\mathbb{E}_{x_r}[\log (1-D(x_{r}))],
\end{aligned}
\end{equation}
% where $z^{\prime} = E(x,v,s)$. 
where $x_{r} = G(z_{r},m_{v_t,s_t})\in \mathbb{R}^{d_x}$ denotes the synthesised tactile features using  random noise $z_{r} \sim \mathcal{N}(0, I)$ and fused features $m_{v_t,s_t}$. 
At the same time, the generator $G$ tries to fool the discriminator, such as by minimising $\mathcal{L}_{G D}^{\prime}=-\mathbb{E}_{\boldsymbol{x_e}}[\log D(x_e)]-\mathbb{E}_{\boldsymbol{x_r}}[\log D(x_r)]$ from a normal GAN's objective~\citep{goodfellow2014generative}.
Through the competition between the discriminator and the generator, the generator is encouraged to synthesise more clear and realistic tactile features.
However, during the training of a GAN, it is found that the real tactile features and the synthesised tactile features are distant between each other, which means that the discriminator can always classify real and synthesised features correctly, i.e., $D(x_t) \rightarrow 1$, $D(x_e) \rightarrow 0$ and $D(x_r) \rightarrow 0$, particularly in the beginning of training.
As a result, it is undesirable for the generator to fool the discriminator and a gradient vanishing problem may occur because of $\partial \mathcal{L}_{G D}^{\prime} / \partial D\left(\boldsymbol{x}_e\right) \rightarrow-\infty$ and $\partial \mathcal{L}_{G D}^{\prime} / \partial D\left(\boldsymbol{x}_r\right) \rightarrow-\infty$.
% where $\partial \mathcal{L}_{G D}^{\prime} / \partial D\left(\boldsymbol{x}^{\prime}\right) \rightarrow-\infty$.
% Consequently, the quality of synthesised tactile features cannot be improved through their competition.
To address this issue, in addition to $\mathcal{L}_{rec}$, we optimise the generator by minimising the mean feature matching loss~\citep{bao2017cvae} between real tactile features and synthesised features:
\begin{equation}\label{eq:4}
\begin{aligned}
\mathcal{L}_{G D}&=\left\|\mathbb{E}_{x_t}[f_{D}({x_t})]-\mathbb{E}_{x_e}[ f_{D}(x_{e})]\right\|_{2}^{2}\\
&+\left\|\mathbb{E}_{x_t}[f_{D}({x_t})]-\mathbb{E}_{x_r}[ f_{D}(x_{r})]\right\|_{2}^{2}.
\end{aligned}
\end{equation}
The centre of synthesised tactile features and the centre of real tactile features should be as close as possible to meet this objective.
It solves the gradient vanishing problem when the synthesised feature and the real feature do not overlap with each other, which allows a more stable training and a faster convergence speed, thus assisting the zero-shot tactile textures recognition.
To summarise, the overall generator loss can be given as:
\begin{equation}
\mathcal{L}_{G}=\lambda_{1}\mathcal{L}_{rec}+\lambda_{2} \mathcal{L}_{G D},
\end{equation}
where $\lambda_{1}$ and $\lambda_{2}$ are hyperparameters weighting the losses of the generator, and are set to $1, 20$  respectively in our experiments through a grid search with a validation set.

For the training process of our proposed generative model, we optimise the encoder $E$, the generator $G$ and the discriminator $D$ iteratively, with the parameters in feature extraction module frozen. 
During training, the visual-semantic fusion module is considered as a component of the generator and updated with the generator as a whole. 
An Adam optimiser~\citep{kingma2014adam} is applied to optimise the model and the learning rates are set to $1\mathrm{e}{-4}$, $1\mathrm{e}{-4}$, $1\mathrm{e}{-5}$ for $E$, $G$ and $D$ respectively.
The training pipeline of our generative model is described in Algorithm~\ref{alg1}.
% To summarize, the overall loss function can be given as:
% \begin{equation}
% \mathcal{L}=\mathcal{L}_{K L}+\lambda_{1}\mathcal{L}_{G}+\mathcal{L}_{D}+\lambda_{2} \mathcal{L}_{G D}+\mathcal{L}_{C}+\lambda_{3} \mathcal{L}_{G C}.
% \end{equation}
% In the above, $L_{KL}$ is related to the encoder which enables a regularized latent space for the generation process.
% $L_{G},L_{GC},L_{GD}$ are loss functions for the generator, which ensure the generator to synthesise the tactile features close to the real features.
% $L_{D}$ and $L_{C}$ are used for identifying the real/synthesised features and the categories of the features respectively. 

%Further improvements could be made by fine-tuning the weights.

\subsection{Tactile Zero-Shot Recognition}\label{recognition}

% \begin{figure}[t]
% 	\centering
% 	\includegraphics[width=1\columnwidth]{classification.pdf}
%     \captionsetup{font={small,stretch=1.25}}
% 	\caption{ \textbf{The recognition framework.} (a) In conventional ZSL, $CLS_u$ is trained using the synthesised untouched tactile features to classify real tactile features.  (b) In GZSL, two classifiers, $CLS_t$ and $CLS_u$, are trained to classify touched tactile features and untouched tactile features, respectively. The input tactile features will be fed into different classifiers according to their novelty detection results.  }
% 	\label{fig:classification}
% % 	\vspace{-1em}
% \end{figure}

\begin{algorithm}[b]
	%\textsl{}\setstretch{1.8}
	\renewcommand{\algorithmicrequire}{\textbf{Input:}}
	\renewcommand{\algorithmicensure}{\textbf{Output:}}
	\caption{Training pipeline of our proposed method}
	\label{alg1}
	\begin{algorithmic}[1]
		\REQUIRE
		The tactile features; the visual features; the semantic embeddings; the number of iterations $K$
		\ENSURE  The parameters of encoder $\phi_{E}$; the parameters of generator $\phi_{G}$; the parameters of discriminator $\phi_{D}$.
        % \ENSURE  the parameters of generator $\theta_{G}$
		
		\FOR{i = 1 to K}
		\STATE Sample $\left\{x_{t}, v_{t}, s_{t},y_{t}\right\}$ from the touched set; 
		\STATE  Sample $z_{e}\sim q\left(z \mid x_t, v_t, s_t; \phi_{E}  \right);$
		\STATE Synthesise tactile features $x_{e} = G(z_{e},m_{v_t,s_t});$
		\STATE Sample a batch of random noise $z_{r} \sim \mathcal{N}(0, I)$;
		\STATE Synthesise tactile features $x_{r} = G(z_{r},m_{v_{t},s_{t}});$
		\STATE $\phi_{E} \gets\phi_{E}-\nabla_{\phi_{E}}\left( \mathcal{L}_{K L}+  \mathcal{L}_{rec}\right)$
		\STATE $\phi_{G} \gets\phi_{G}-\nabla_{\phi_{G}}{L}_{G}$ (The visual-semantic fusion module is considered as a part of the generator)
		\STATE $\phi_{D}\gets\phi_{D}-\nabla_{\phi_{D}}\mathcal{L}_{D}$
% 		\STATE $\phi_{C}\gets\phi_{C}-\nabla_{\phi_{C}}\mathcal{L}_{C}$
% 		\STATE$i \gets i+1$
		\ENDFOR
% 		\State \textbf{end}
        
	\end{algorithmic}  
% \vspace{-1em}
\end{algorithm}

After training the generative model using touched materials, we can use the generator to synthesise the tactile features of untouched materials, with semantic features and visual features.
The synthesised tactile features can be represented as $x_{syn} = G(z_{r},m_{v_{u},s_{u}})$, where $z_{r} \sim \mathcal{N}(0, I)$.

In the conventional ZSL, a Softmax classifier $CLS_{u}$ is trained using the synthesised tactile features of untouched materials. The classifier minimises the following loss:
\begin{equation}
L_{cls_{u}}=-\frac{1}{|\mathcal{T}_{syn}|} \sum_{(x_{syn}, y_{u}) \in \mathcal{T}_{syn}} \log \left(p\left(y_{u} \mid x_{syn} ; \phi_{u}\right)\right)
\end{equation}
where $\phi_{u}$ is the  parameters of the classifier $CLS_{u}$, and $\mathcal{T}_{syn} =\left\{(x_{syn},y_{u})\right\} $. 
Then, we can use the learnt classifier to classify the tactile features of untouched materials.
% as shown in Fig.~\ref{fig:classification} (a).
The label of the test data can be predicted by: 
\begin{equation}
\hat{y}=\arg \max _{y \in Y_{u}} p\left(y \mid x_{u} ; \phi_{u}\right).
\end{equation}

In GZSL, the key is to understand if the input is from touched classes or untouched classes. 
% \guan{
% Our data collection process (more details in Section~\ref{datacollection}) involves 50 pieces of fabric where each piece is touched by the tactile sensor repeatedly by multiple times, and the data distribution satisfies Central Limit Theorem.}
% To tackle this problem, we use a Gaussian distribution for novelty classification.
Considering the mathematical simplicity and tractability, we apply a Gaussian distribution to model the data and measure the distribution of probability density of touched tactile features~\citep{socher2013zero}. 
An input sample is categorised as touched if it locates in a high-density region, and as untouched if it does not.
Concretely, we use the touched set $\left\{x_{t}^{1}, x_{t}^{2}, \ldots, x_{t}^{n}\right\}$ to determine the parameters $\phi_{gau} = (\mu, \sigma^{2})$ of the distribution by maximum likelihood estimation:
\begin{equation}
\hat{\phi}_{gau}=\arg \max _{\phi_{gau}} \log \prod_{i=1}^{n} p\left(x_{t}^{i} ; \phi_{gau}\right),
\end{equation}
then if $\log p(x ; \phi_{gau})$ is greater than a selected threshold $\beta$, $x$ is from the touched classes, otherwise, it is from the untouched classes. 
% The threshold $\beta$ is tuned with the validation set that is able to maximize the mean accuracy of classifying $x_{train}\in D_{train}$ as touched and $x_{val}\in D_{val}$ as untouched. 

Apart from the classifier $CLS_{u}$, we train another Softmax classifier $CLS_{t}$ with touched set to recognise the data from touched materials.
Accordingly, the test tactile features can be fed into different classifiers for recognition.
% as shown in Fig.~\ref{fig:classification} (b). 
The label of the test data can be predicted by:
\begin{equation}    
\hat{y}=    \begin{cases}\arg \underset{y\in Y_{t}}\max\quad p\left(y \mid x ; \phi_{t}\right)&\mbox{if $ \log p\left(x ; \phi_{gau}\right) > \beta$  }\\    
\arg \underset{y\in Y_{u}}\max\quad p\left(y \mid x ; \phi_{u}\right)&\mbox{otherwise.}    \end{cases}   \end{equation}

\subsection{Network Implementation}

%In this work, we apply pretrained models to extract high dimensional features to represent visual images, tactile textures, and semantic attributes.

Table~\ref{tbl:structure} demonstrates the network structures of different components.
The encoder $E$ consists of three Fully Connected (FC) layers, where the second and the third layers are both connected to the first layer to produce the mean vector and the variance vector respectively, and each layer is followed by a LeakyReLu activation function.
The visual-semantic fusion module includes three FC layers where the first two layers are each followed by a ReLU activation function, and the last layer is followed by a Sigmoid activation function.
The generator $G$ is a network with two FC layers, and each layer is followed by a LeakyReLU activation function. 
% \guan{The encoder E will be described with more details}
For the discriminator $D$, we use a network with two FC layers followed by a LeakyReLU and a Sigmoid activation functions respectively for binary classification.
The networks of $CLS_{t}$ and $CLS_{u}$ are both implemented by three FC layers, where first two layers are each followed by a LeakyReLU activation function, and the last layer is followed by a Softmax activation function. 
The hyperparameters, such as the number of nodes, are tuned by using the validation set so as to achieve a balance between the complexity of the model and the ZSL performance.

All of the networks in our framework are implemented by the Keras using TensorFlow backend with Python. 
The scikit-learn is also implemented to calculate the parameters of the Gaussian distribution.
Our models are trained on a PC with an AMD Ryzen 7 3700X 8-Core Processor, an Nvidia 2080Ti graphic card and 16 GB RAM.

% \begin{figure}
% 	\centering
% 	\includegraphics[width=1\columnwidth]{attention.pdf}
%     \captionsetup{font={small,stretch=1.25}}
% 	\caption{ \textbf{The framework of multimodal fusion function.} $v$ represents the visual feature; $s$ represents the semantic feature; $\otimes$ represents element-wise product.  }
% 	\label{fig:attention}
% % 	\vspace{-1em}
% \end{figure}

\begin{table}[h]
% \vspace{-1.5em}
\centering
% \captionsetup{font={small,stretch=1.25}}
\caption{\textbf{Network implementation in our framework.}}
\small
%\adjustbox{max width=\linewidth}{%
\begin{tabular}{c|c|cc}

\hline
   \multirow{2}{*}{Network} & \multirow{2}{*}{Layers} & \multicolumn{2}{c}{Parameters} \\
 \cline{3-4}
  &  & Number of layers & Number of Nodes  \\
\hline
$E$  & $FC$& $3$&$2048-2048-2048$ \\
$VS-F$  & $FC$& $3$&$512-256-2072$ \\
% \hline
$G$ &$FC$ &$2$ & $2048-2048$\\
% \hline
$D$ &$FC$ &$2$ & $512-1$\\
% \hline
% \hline
$CLS_{t}$ &$FC$ &$3$ & $512-512-40$\\
% \hline
$CLS_{u}$ &$FC$ &$3$ & $512-512-5$\\
\hline
\end{tabular}

\label{tbl:structure}
% \vspace{-0.5em}
%\vspace{-5mm}
\end{table}

\section{Experimental Setup}\label{datacollection}

\begin{figure*}[t]
	\centering
	\includegraphics[width=2\columnwidth]{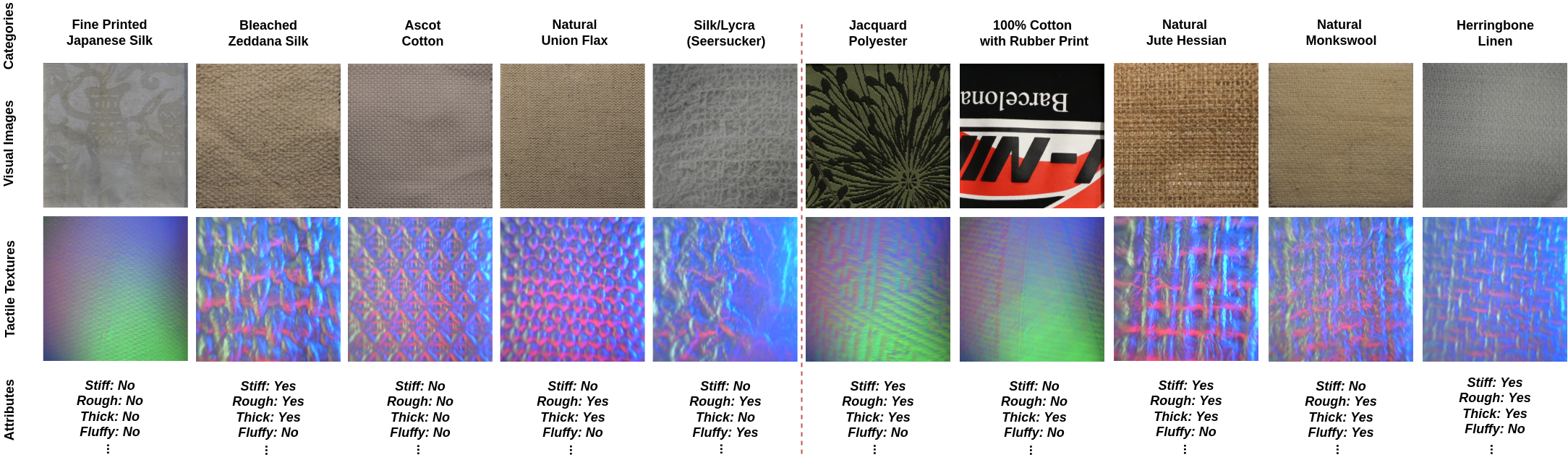}
% 	\captionsetup{font={small,stretch=1.25}}
	\caption{\textbf{Examples in the dataset.} The first row: categories of different fabrics. The second row: visual images recorded by a digital camera. The third row: tactile textures collected from fabrics by a GelSight sensor. The fourth row: semantic attributes measured by human observations. Left five columns: samples of five example training classes from the training set (we have 40 different fabrics in the training set in total). Right five columns: samples from the test classes (we have 5 unknown fabrics for the test in total). }
% 	In the collection of data, a digital camera is used to take the visual image of the fabric, whereas a GelSight sensor, mounted on a UR5 robot arm, is pressed against the fabric to collect the tactile textures.
% 	The locations of the pressing are recorded, and each tactile texture can be paired with a certain location on the visual image, e.g., the tactile texture in the first column is corr
	\label{fig:example}
	\vspace{-1em}
	
\end{figure*}

\begin{figure}
	\centering
	\includegraphics[width=0.9\columnwidth]{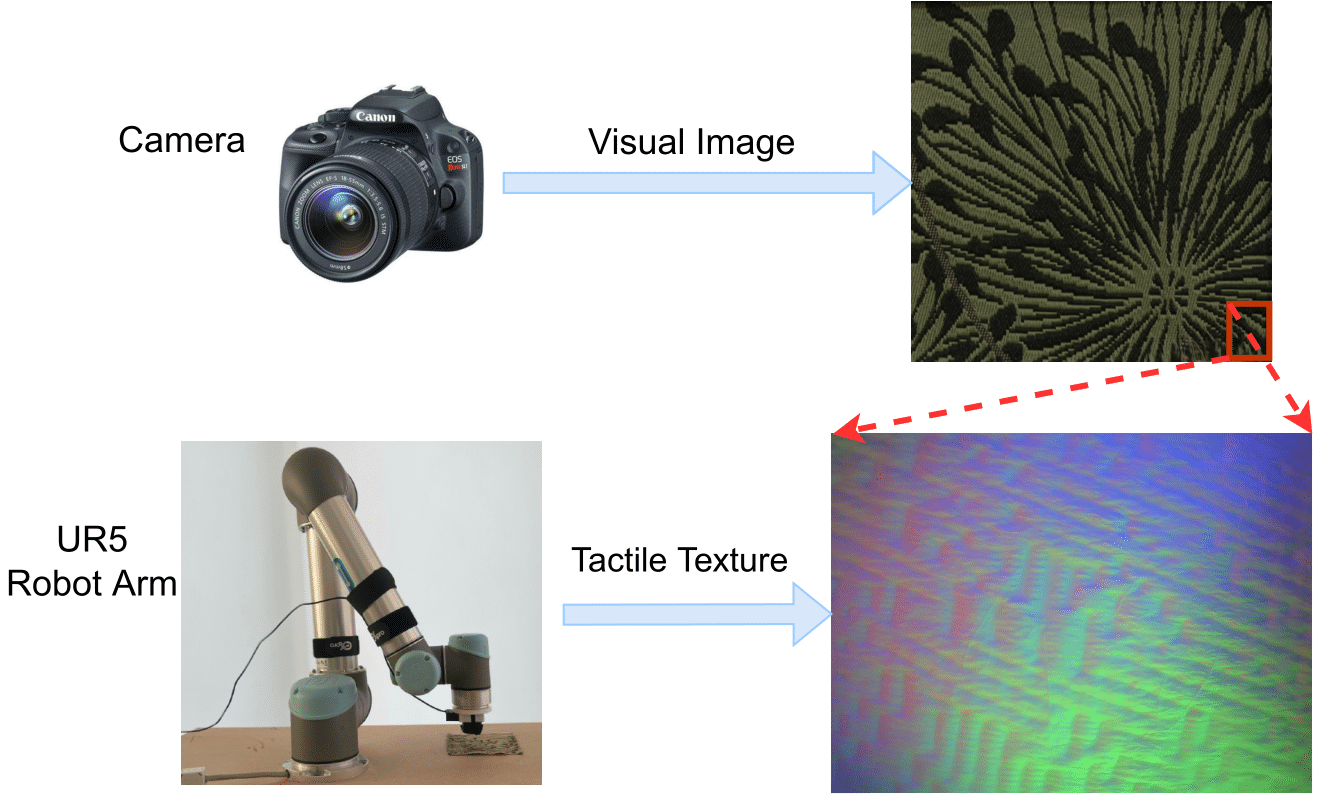}
    % \captionsetup{font={small,stretch=1.25}}
	\caption{\textbf{Illustration of data collection. } A digital camera is used to take the visual image of the fabric, whereas a GelSight sensor, mounted on the UR5 robot arm, is pressed against the fabric to collect the tactile textures.
	The pressing locations are recorded, and each tactile texture can be paired with a certain location on the visual image. 
	The example texture is corresponding to the red rectangle region of the visual image.}
	\label{fig:data}
	\vspace{-1.5em}
\end{figure}

\noindent \textbf{Fabric materials.} 
% Our dataset includes visual images, tactile textures, and semantic attributes from 50 different fabrics, and some examples are illustrated in Fig.~\ref{fig:example}. 
There are 50 different fabrics used in our experiments.
% 50 different fabrics are selected for data collection, including the cotton, silk, polyester, linen, etc., and contain some mixture of different materials, such as wool and cotton mixture.
The fabrics include pure materials like cotton, silk, polyester, linen, etc., as well as some mixtures of different materials, such as cotton mixed with wool.
% About $30\%$ of the fabrics have different color patterns with embroidery, jacquard or pigment paint, and others are of single color.
% Approximately $30\%$ of the fabrics have different colored patterns while others are made from a single color.
Around $30\%$ of the fabrics have different coloured patterns while the rest come in single colours.
Most of the fabrics are cloth pieces of  $8 cm \times 8 cm$, whereas some of the fabrics are from daily clothing, and an $8 cm \times 8 cm$ area from each daily clothing is selected to collect the data.
In addition, the fabrics are similar in thickness, ranging from 1mm to 2mm.
Specifically, the fabrics are divided randomly with a ratio of 8:1:1, i.e., 40 fabrics are used to collect the training set $D_{t}$; 5 fabrics are used to collect the validation set $D_{val}$ to tune the model; and 5 fabrics are used as untouched materials for testing.
Some examples are shown in Fig.~\ref{fig:example}.

\noindent \textbf{Data acquisition for training.} 
\begin{table}
	\centering
		\caption{\textbf{Comparison with other dataset.} We compare our collected dataset with other datasets with triple modalities. }
		\label{tab:dataset}
        \scalebox{0.90}{
		\begin{tabular}{c| c | c | c}
			\hline
			\multirow{2}{*}{Dataset} & \multirow{2}{*}{{\makecell[c]{PHAC-2~\citep{chu2015robotic}}}} & \multirow{2}{*}{{\makecell[c]{~\citep{yuan2017connecting}}}}&
			\multirow{2}{*}{{\makecell[c]{Ours}}}
            \\&\multicolumn{1}{c|}{}
            &\multicolumn{1}{c|}{}\\
			\hhline{=|=|=|=}
        \multirow{2}{*}{Objects} & \multirow{2}{*}{\makecell[c]{Household items\\/Raw materials}} & \multirow{2}{*}{Fabrics}&
			\multirow{2}{*}{Fabrics}
            \\&\multicolumn{1}{c|}{}
            &\multicolumn{1}{c|}{}\\
		\hline
        
		Tactile sensor & BioTac &GelSight&GelSight  \\
		\hline
		Number of objects & 60 &118&50  \\
        \hline

		Number of attributes & 24 &4&24  \\
		\hline
		
		\multirow{2}{*}{\makecell[c]{Number of trials\\per object}} & \multirow{2}{*}{{10}} & \multirow{2}{*}{{25}}&
			\multirow{2}{*}{{225}}
            \\&\multicolumn{1}{c|}{}
            &\multicolumn{1}{c|}{}\\
		\hline
		\end{tabular}}
\end{table}
To train and tune the model with multimodal data, visual images, tactile textures and semantic attributes are collected from fabrics that are used for training and validation.
The test fabrics are only visually inspected and semantically annotated without collecting tactile data.
To collect visual data, we use a digital camera Canon 4000D to record the target areas from a fixed distance of 30$cm$.
The fabric is laid flat on a horizontal plane, while the image plane is parallel with the fabric as well. 
As shown in the visual images of Fig.~\ref{fig:example}, we crop the main body of the fabric, and remove the extra background. The resolution after cropping is $1000\times1000$.

As shown in Fig.~\ref{fig:data}, a GelSight sensor is applied to record tactile textures by pressing against the fabrics. 
To collect the data automatically, a Universal Robots UR5 is applied to collect data using Robot Operating System (ROS).
Different from the visual images taken from a distance, the tactile texture is captured only locally by physical contact.
As a result, there exists a scale gap between vision and tactile sensing.
To reduce the large scale gap between the visual and tactile sensing, we expect that each tactile texture has a paired visual image at the same location of the fabric.
Specifically, the robot arm is controlled to press against fabrics by about 15N with the GelSight sensor that has around $1.5cm\times1.1cm$ perception field with a resolution of $640\times480$, starting from a corner, moving along the horizontal and vertical directions with a step length of $4.9mm$ and $4.6mm$ respectively, until covering an $8 cm \times 8 cm$ area.
As a result, each tactile texture in our dataset is corresponding to a certain location in a visual image, reducing the scale gap significantly.
Finally, each fabric is contacted by the sensor 225 times for tactile data collection.
To pair with each tactile texture, each visual image is cropped into 225 parts according to the contact locations as well.

To collect semantic attributes, each fabric is labelled by humans according to its physical characteristics through visual observation, including \textit{{stiff, soft, rough, smooth, thick, thin, cool, warm, fluffy, heavy, delicate, durable, stretchable, absorbent, holey, flat, bumpy, patterned, striped, shiny, hairy, embroidered, jacquard, pigment printed}}, to characterise the high-level features of fabrics~\citep{venkatraman2015fabric}.
Each attribute was given a \textit{True} or \textit{False} value according to their properties.
The first six paired attributes are exclusive, and only one attribute can be given \textit{True} from each pair.
For example, if \textit{stiff} is given \textit{True} value, attribute \textit{soft} must be \textit{False}.
The attributes are measured by 5 researchers who work on tactile sensing, and we compute the final attribute values by majority voting. 
Then, we use the one-hot encoding to get the semantic representation.
For example, if the attributes of one piece of fabric is \textit{\{True, False, False, ..., True, True\}}, the semantic vector would be [1, 0, 0, ..., 1, 1].
Though our fabric semantic attributes may not encompass all aspects of fabric characteristics and are subject to human bias, they can still give us valid and appropriate information to describe the properties of fabrics.

By comparing our dataset against other tactile datasets (as shown in Table~\ref{tab:dataset}), there are several advantages of our dataset: 
% (1) the tactile data is collected by a sensitive GelSight sensor, and the tactile images can provide us various physical tactile properties for representation; 
(1) compared with the datasets in~\citep{chu2015robotic,yuan2017connecting}, the scale gap between the visual data and the tactile data is reduced significantly as the visual image is cropped according to the contact location; 
(2) a UR5 robot arm is applied to collect the data automatically, which is more stable and alleviates human error compared to collecting data manually in~\citep{yuan2017connecting};  
% (3) our dataset contains 11,250 weakly paired visual and tactile images with corresponding semantic attributes by 225 explorations on each object of different locations
(3) each object is explored by the tactile sensor for 225 times, which results in a larger dataset than existing datasets~\citep{chu2015robotic,yuan2017connecting} with triple modalities (i.e., vision, touch and semantic attributes).

\noindent \textbf{Different information embeddings of visual data and tactile data.} 
The visual data are captured from a distance by the digital camera where the global information such as appearance, shapes and colours of materials will be recorded. 
Different from visual data, tactile data are collected by physical contact between objects and an optical tactile sensor, i.e., the GelSight sensor.   
When the GelSight sensor interacts with the objects, the elastomer on the sensor will be deformed in response to the contact force, and the surface geometry will be mapped into the deformation, which is captured by the camera inside the sensor.
% Through the reflective membrane coated on the elastomer, the camera inside the sensor is able to capture the deformation.
In~\citep{yuan2016estimating,yuan2017gelsight}, it is demonstrated that the tactile images from GelSight have the ability to indicate tactile textures, contact force, surface height, hardness, etc.
Though the visual data and tactile data are of the same format (i.e., images), they reflect different properties of the objects.

% Our dataset will be available upon acceptance at  https://sites.google.com/view/multimodalzsl.

\noindent \textbf{Tactile zero-shot recognition task.} 
After training the model with the training set  $D_{t}$ and validation set $D_{val}$, we enable the robot to press against the untouched test materials to collect tactile textures with the GelSight sensor for zero-shot recognition.
Concretely, each test material is pressed by the sensor to collect tactile textures 225 times at different locations, with the same tactile data sampling method described in data acquisition, which are then recognised by the trained model. 
Some robotic experiment demos are shown on our website\footnote{https://sites.google.com/view/multimodalzsl}.

\section{Experimental Results and Analysis}\label{experiments}
In this section, a series of experiments are conducted to evaluate our proposed VS2T-ZSL method for the tactile ZSL problem.
The goal of the experiments is three-fold: 
1) To learn how different components in our proposed structure improve the ZSL results;
2) To investigate the synergistic effects of multimodal input;
3) To evaluate the effectiveness of our proposed method against other methods in both ZSL and GZSL settings.

\subsection{Results of VS2T-ZSL}

\begin{table}
	\centering
% 		\caption{Texture recognition results using different models when different number of tactile images are used in a sequence.}
        % \captionsetup{font={small,stretch=1.25}}
		\caption{Recognition accuracy of untouched materials using various network structures. }
		\label{ablation1}
        \scalebox{0.85}{
		\begin{tabular}{cccc| c | c | c }
			\hline
			\multicolumn{4}{c|}{Network structure}&
			\multirow{2}{*}{{\makecell[c]{Average\\Accuracy $\uparrow$}}}&
			\multirow{2}{*}{{\makecell[c]{Warsserstein\\Distance $\downarrow$}}}&
			\multirow{2}{*}{{\makecell[c]{Cosine\\Similarity $\uparrow$}}}\\
			\cline{1-4}
            % E&VS-F &G^* &D^*
            E&VS-F &$G^*$ &$D^*$
            &\multicolumn{1}{c|}{}
            &\multicolumn{1}{c|}{}
            &\multicolumn{1}{c}{}\\
			\hhline{====|=|=|=}
		
		&&$\checkmark$&$\checkmark$ &  $74.61\%$&$1788$&$0.41$  \\
		&$\checkmark$&$\checkmark$&$\checkmark$ & $78.21\%$ &$1571$&$0.43$  \\
		$\checkmark$&&$\checkmark$& $\checkmark$ & $79.50\%$ &$1320$&$0.47$  \\
		$\checkmark$&$\checkmark$&$\checkmark$&$\checkmark$ & \pmb{$83.06\%$} &\pmb{$1165$}&\pmb{$0.51$} \\
		\hline
		\end{tabular}}
\end{table}

\begin{table}
	\centering
% 		\caption{Texture recognition results using different models when different number of tactile images are used in a sequence.}
        % \captionsetup{font={small,stretch=1.25}}
		\caption{Recognition accuracy of untouched materials using different input modalities. }
		\label{tab:unimodal}
        \scalebox{0.90}{
		\begin{tabular}{c| c | c | c}
			\hline
			\multirow{2}{*}{Input Modality} & \multirow{2}{*}{{\makecell[c]{Average\\Accuracy $\uparrow$}}} & \multirow{2}{*}{{\makecell[c]{Warsserstein\\Distance $\downarrow$}}}&
			\multirow{2}{*}{{\makecell[c]{Cosine\\Similarity $\uparrow$}}}
            \\&\multicolumn{1}{c|}{}
            &\multicolumn{1}{c|}{}\\
			\hhline{=|=|=|=}
% 			\hline
        % vision_a & 84.18\%/78.24\% & 96.1\%/85.4\% & 11.11\% & 11.11\% & 11.11\% & 11.11\%\\
		S2T-ZSL & $55.07\%$ &1458&0.31  \\
		V2T-ZSL & $72.71\%$ &1364&0.46  \\
		VS2T-ZSL & $\pmb{83.06\%}$ &\pmb{1165}&\pmb{0.51}  \\
		\hline
		\end{tabular}}
  \vspace{-1em}
\end{table}

To investigate how different components work in synthesising tactile features in ZSL, we conduct an ablation study for our proposed VS2T-ZSL structure. 
Due to the fact that $G$ and $D$ are necessary components to synthesise tactile features, we explore the effect of removing $E$ and VS-F individually, as well as removing them together.
To ensure that our results are determined by the proposed method rather than the chosen objects, we repeat the aforementioned process of dividing materials randomly for five times and calculate the average results over all splits to validate the robustness of our proposed method. 
% For example, ``w/o discriminator" indicates that the method is implemented without the discriminator.

As shown in Table~\ref{ablation1}, the recognition accuracy of untouched materials is given to evaluate the performance in ZSL.
Our proposed VS2T-ZSL is able to achieve a higher recognition accuracy of $83.06\%$ to recognise the unknown materials, compared to the results when a certain component is removed.  
In particular, there is an obvious drop by $4.85\%$ when $E$ is removed, which demonstrates that the continuous latent space generated by the encoder makes a great contribution to synthesising untouched tactile features.
The absence of the VS-F produces inferior recognition results, decreased by $3.56\%$, compared to the result of VS2T-ZSL. 
This indicates that the visual-semantic fusion module is able to select the salient features across different modalities for the generation task. 
% The encoder network enables a continuous space for generation process, and the absence of the encoder decreases the recognition accuracy by $10.0\%$.
% Furthermore, $16\%$ recognition accuracy increase can be obtained by using multimodal fusion method, which emphasizes the salient features across different modalities for generation.
Moreover, there is a decrease of $8.45\%$ when both $E$ and VS-F are removed, compared to the result of VS2T-ZSL. 

Furthermore, the Wasserstein distance is implemented to measure the distance between the distributions of the real tactile features and synthesised features of untouched materials. 
As shown in Table~\ref{ablation1}, it is observed that the Wasserstein distance between synthesised tactile features and real tactile features is 1,165 by using the VS2T-ZSL, which are hundreds less than the results from the ablated structures.
It means that the distribution of the generated features with the VS2T-ZSL method are closest to the real distribution.

The cosine similarity is also given to evaluate the similarity between each synthesised tactile features and real tactile features, and the average scores are shown in Table~\ref{ablation1}. It can be seen that the proposed VS2T-ZSL achieves the highest similarity score, i.e., 0.51, among all the structures, which demonstrates that the synthesised features are more similar to the real features by using our proposed structures.
\begin{figure}[t]
	\centering
	\includegraphics[width=1\columnwidth]{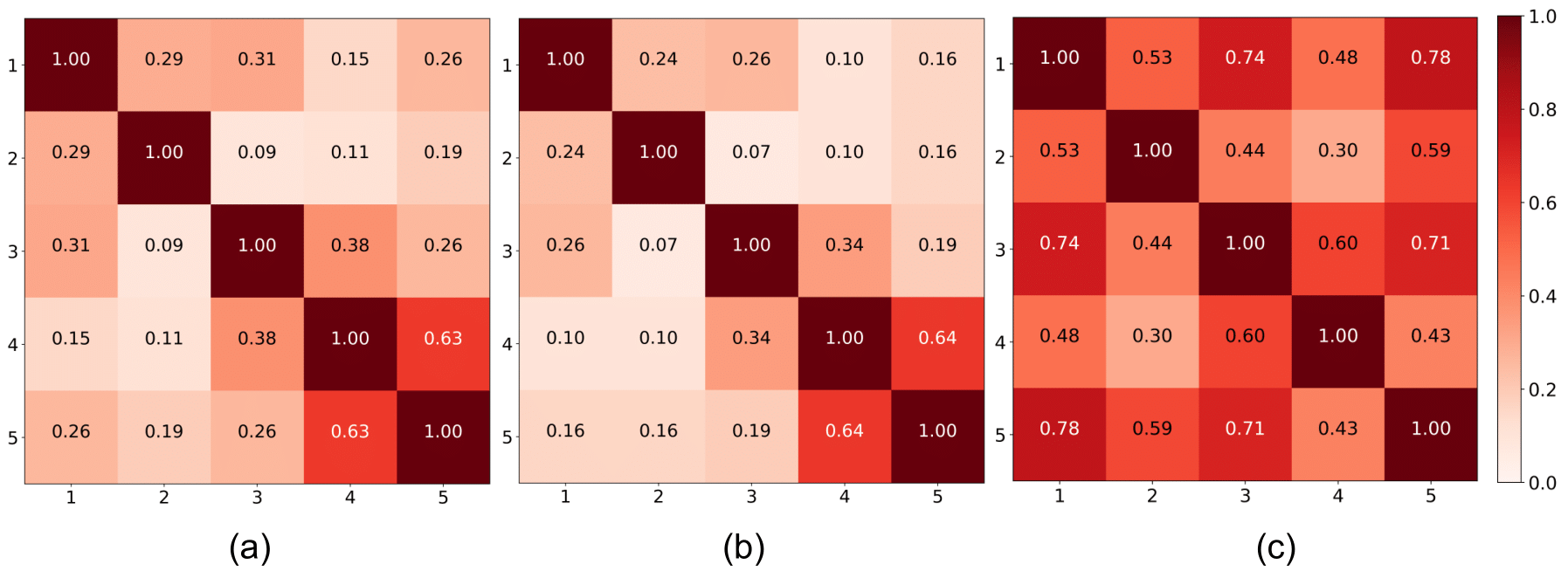}
    % \captionsetup{font={small,stretch=1.25}}
	\caption{\textbf{Cosine similarities}: (a) similarity of multimodal input between each class, (b) similarity of visual input between each class, (c) similarity of semantic input between each class.}
	\label{fig:sim}
	% \vspace{-1.em}
\end{figure}
\begin{figure}[t]
	\centering
	\includegraphics[width=1\columnwidth]{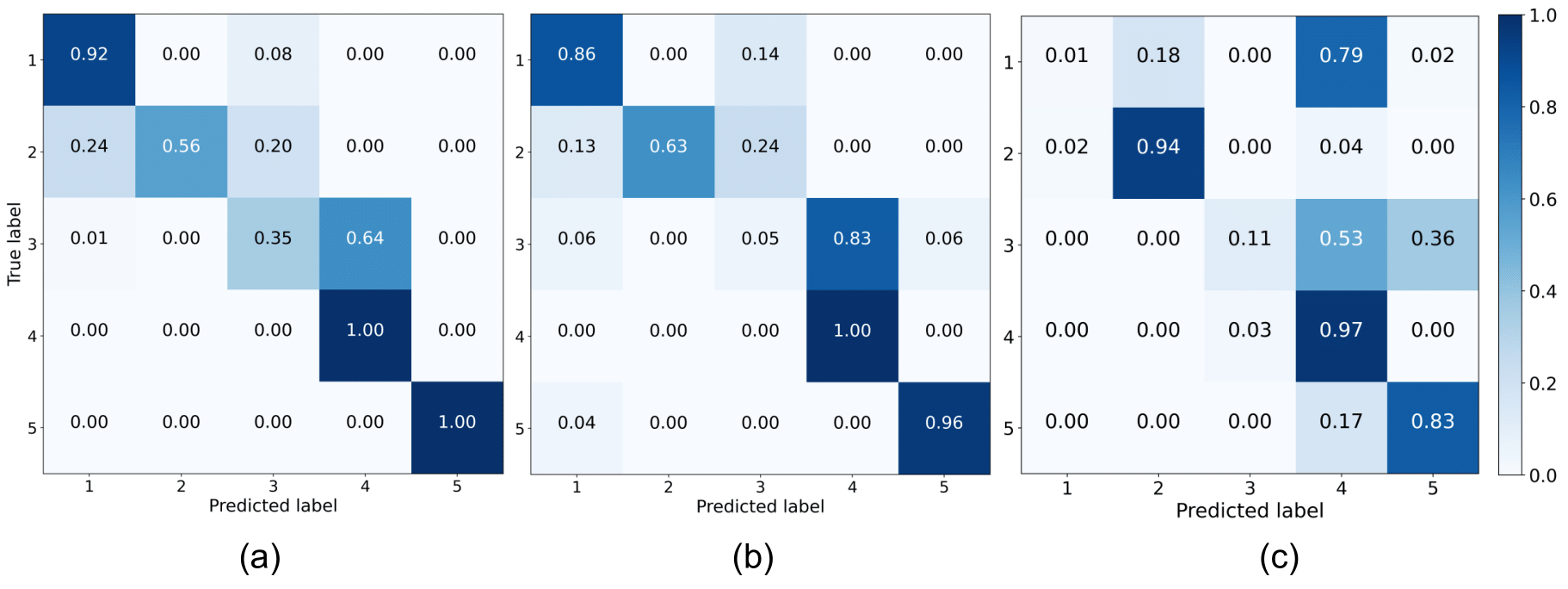}
    % \captionsetup{font={small,stretch=1.25}}
	\caption{\textbf{Normalised confusion matrices}: (a) results of the model with multimodal input; (b) results of the model with the visual input; (c) results of the model with the  semantic input. }
	\label{fig:conf}
	\vspace{-1.2em}
\end{figure}
It can be concluded that our proposed VS2T-ZSL method, which consists of a feature extraction module, an encoder, a visual-semantic fusion module, a generator and a discriminator, allows the generated features of untouched material to be more realistic, thus providing a better performance in tactile ZSL.

% In addition, it is highly recommended that reviewers check out the robotic experiment demo on our website. In the demo, a GelSight sensor mounted onto a robot arm is used to identify materials that have never been touched before, using our proposed model. Since our paper focuses on designing the algorithm of tactile ZSL, the demo is not included in the paper and is only intended to demonstrate the potential application.

\begin{figure}
    % \vspace{-1em}
	\centering
	\includegraphics[width=0.8\columnwidth]{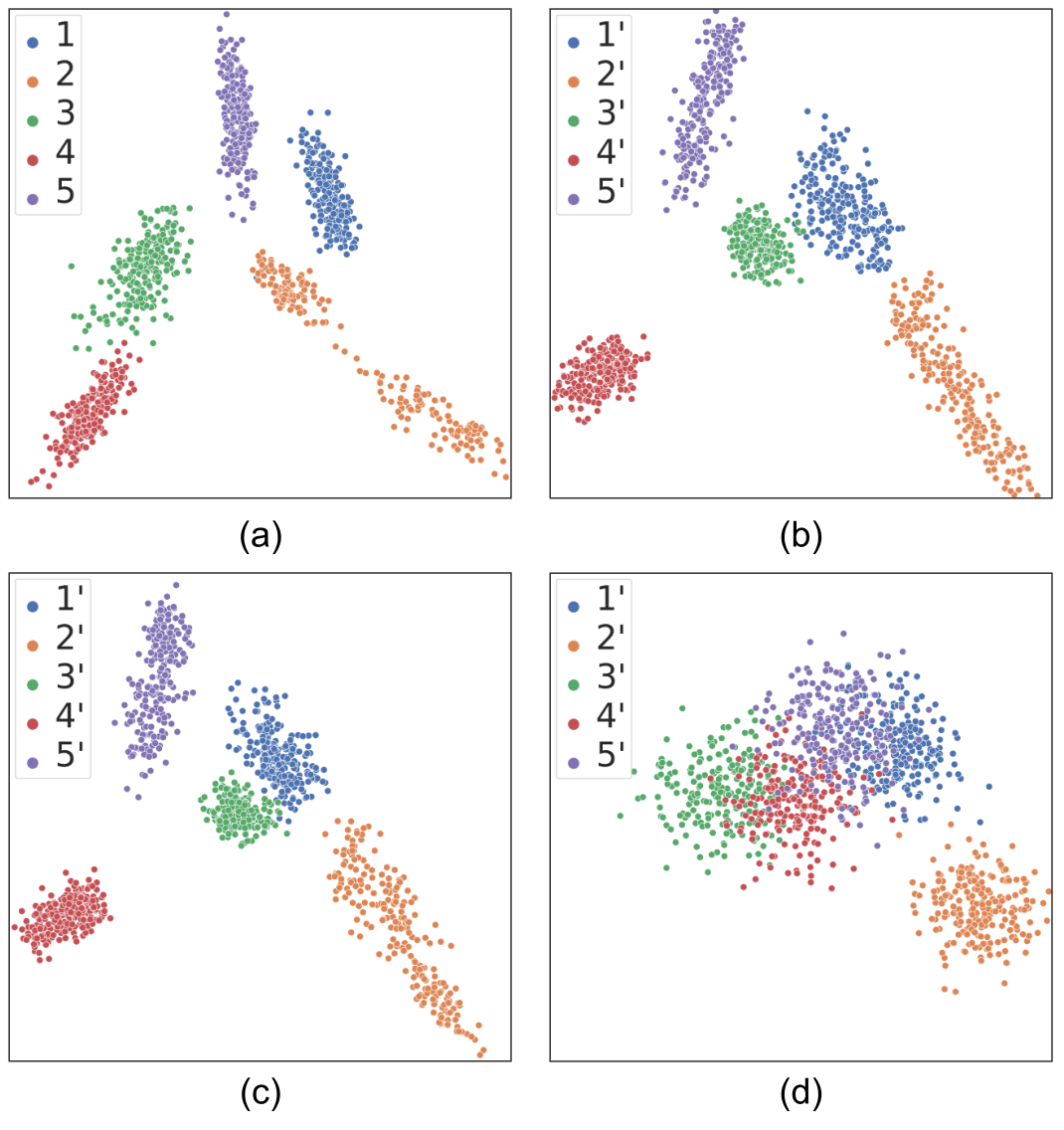}
% 	\captionsetup{font={small,stretch=1.25}}
	\caption{\textbf{Tactile feature distributions.} (a) real tactile features of untouched materials; (b) synthesised tactile features with multimodal input; (c) synthesised tactile features with visual input; (d) synthesised tactile features with semantic input.}
	\label{fig:distribution}
	\vspace{-1.5em}
\end{figure}

% \begin{figure}[t]
% 	\centering
% 	\includegraphics[width=0.1\columnwidth]{proto.pdf}
%     % \captionsetup{font={small,stretch=1.25}}
% 	\caption{\textbf{Projected input features and ground-truth semantic embeddings.} (a) projected tactile features (triangle markers) and ground-truth semantic embeddings (star markers) of untouched materials with DAP. (b) Projected visual and tactile features (square markers) and ground-truth semantic embeddings (star markers) with VT-FC-ZSL.}
% 	\label{fig:shift}
% % 	\vspace{-1em}
% \end{figure}

\begin{table*}[]
\centering
% \small
% \adjustbox{max width=\linewidth}
% \resizebox{\linewidth}{!}
% \captionsetup{font={small,stretch=1.25}}
\caption{\textbf{Comparison results.} We compare our results against other methods in both ZSL and GZSL settings. $acc_t$ and $acc_u$ represent the accuracy of touched and untouched materials, respectively; $H=\frac{2 * acc_u * acc_t}{acc_u + acc_t}$ denotes their harmonic mean. }
\scalebox{1}{
{\begin{tabular}{c|c|c|c|c|c|ccc}
\hline
   \multirow{2}{*}{Methods} & \multirow{2}{*}{{\makecell[c]{Gene-\\rative}}}& \multirow{2}{*}{{\makecell[c]{Projection\\based}}}& \multirow{2}{*}{{\makecell[c]{Uni-\\modal}}}& \multirow{2}{*}{{\makecell[c]{Multi-\\modal}}}& \multirow{2}{*}{\makecell[c]{ZSL \\ $acc $ $\uparrow$ }} & \multicolumn{3}{c}{GZSL} \\
 \cline{7-9}
  & & & &&&$acc_{t}$ $\uparrow$ & $acc_{u}$ $\uparrow$ & $H$ $\uparrow$ \\
% \hline
\hhline{=|=|=|=|=|=|===}
DAP~\citep{abderrahmane2018haptic}&  &$\checkmark$ &$\checkmark$ & &$43.56\%$ &$94.11\%$ & $43.47\%$&$59.47\%$ \\
VT-FC-ZSL~\citep{abderrahmane2018visuo}&  &$\checkmark$ & &$\checkmark$ &$63.20\%$ &$94.11\%$ & $62.58\%$&$75.17\%$ \\
Abderrahmane \textit{et al.}~\citep{abderrahmane2019deep}& $\checkmark$ & &$\checkmark$ & &$47.20\%$ &$83.44\%$ & $47.11\%$&$60.22\%$ \\
Lee \textit{et al.}~\citep{lee2019touching}&$\checkmark$ & &$\checkmark$ & & $51.29\%$&$89.00\%$ &$51.20\%$ & $65.00\%$\\
VS2T-ZSL&$\checkmark$ & & &$\checkmark$ &\pmb{$76.36\%$} &\pmb{$94.56\%$} & \pmb{$76.00\%$}&\pmb{$84.27\%$}\\
\hline
\end{tabular}}}

\label{tab:compare}
%\vspace{-5mm}
\vspace{-1em}
\end{table*}

\subsection{Multimodal vs. Unimodal Input}

% \begin{figure}[t]
% 	\centering
% 	\includegraphics[width=1\columnwidth]{confusion.pdf}

% 	\caption{Normalized confusion matrices with (a) multimodal input, (b) visual input, and (c) semantic input, respectively.}
% 	\label{fig:confusion}
% % 	\vspace{-1em}
% \end{figure}

% \begin{figure}[t]
% 	\centering
% 	\includegraphics[width=1\columnwidth]{score.pdf}

% 	\caption{Cosine similarities between different classes, with (a) multimodal input, (b) visual input, and (c) semantic attributes, respectively.}
% 	\label{fig:similarity}
% % 	\vspace{-1.5em}
% \end{figure}

To investigate the synergistic effect of multimodal input, we compare our VS2T-ZSL with the methods using unimodal inputs.
Firstly, we only apply semantic attributes to generate the tactile features of untouched materials in ZSL (S2T-ZSL).
Secondly, we only use the visual images to synthesise the tactile features (V2T-ZSL).

As shown in Table~\ref{tab:unimodal}, the recognition accuracy is only $55.07\%$ from S2T-ZSL, while it is $72.71\%$ from V2T-ZSL, which are much lower than the $83.06\%$ recognition accuracy from VS2T-ZSL.
The Wasserstein distance and the cosine similarity also show that the generated features with multimodal input are much closer to the real features than with unimodal inputs.
The results indicate that multimodal input enables us to produce more realistic tactile features from different perspectives with multiple sources, which is beneficial to zero-shot recognition.
% It indicates the multimodal input enables us to generate more realistic tactile features with different perspectives from multiple domains, which is beneficial to zero-shot recognition.

Particularly, it can be observed that the recognition accuracy has an obvious drop, by $17.64\%$ and $27.99\%$ respectively, while using semantic attributes only compared to the results by using visual input only or multimodal input.
A possible reason causes this difference is due to the limited number of semantic attributes used in our approach.
There are only 24 attributes used to describe fabric characteristics. 
The semantic attributes cannot encompass all of the fabric's characteristics and may have many overlapping values between different materials.
As shown in Fig.~\ref{fig:sim}, we take the data from the first materials split as an example and compute the cosine similarity between the mean values of the input features of each untouched class for multimodal input and unimodal input respectively.
It can be seen that the semantic input of each class is more similar to each other compared to the visual input and multimodal input.
Since the input of each class is similar, it will be difficult to generate recognisable tactile features from semantic information, which results in a lower recognition accuracy in ZSL.
Compared with semantic attributes, visual input and multimodal input (as can be seen in Fig.~\ref{fig:sim} (a) (b)) have a smaller similarity between each class, which enables us to synthesise discriminative tactile features for tactile ZSL tasks.

To further analyse the synergistic effect of the multimodal input, we illustrate the normalised confusion matrices using different input modalities in Fig.~\ref{fig:conf}.
% It can be found that the combination of visual and semantic information enables a synergistic effect where the result cannot be achieved with a single modality.
Specifically, we can see that only $5\%$ and $11\%$ tactile features of class 3 (Natural Jute Hessian, see the 8th column in Fig.~\ref{fig:example}) are classified correctly using the visual input and semantic input respectively (as shown in Fig.~\ref{fig:conf} (b) (c)), whereas $35\%$ tactile features are classified correctly with the multimodal input (as shown in Fig.~\ref{fig:conf} (a)).
It indicates that by combining visual and semantic information, we can represent the characteristics of materials from different domains and mitigate biases in a single domain, resulting in a synergistic effect where the result cannot be achieved with a single modality.  

What is more, it is worth noting that most tactile data of class 3 (Natural Jute Hessian) are misclassified with both multimodal input and unimodal input.
To better understand the reason, we analyse it in terms of both the output distribution and the input similarity.
% As shown in Fig.~\ref{fig:sim}, we compute the cosine similarity between the mean values of the input features of each class for multimodal input and unimodal input respectively.
The distributions of real tactile features and synthesised tactile features of untouched materials are shown in Fig.~\ref{fig:distribution}, by using Principal Component Analysis (PCA) where we reduce the data dimension to visualise high dimensional tactile features in 2D.
If the classifier $CLS_u$ is trained with the synthesised features (e.g., features in Fig.~\ref{fig:distribution} (b)) and is used to predict the categories of real tactile features (in Fig.~\ref{fig:distribution} (a)), the testing tactile features of class 3 (Natural Jute Hessain marked in green) have a higher probability to be misclassified to a closer cluster in synthesised features of class 4 (Natural Monkswool marked in red) due to the bias in the synthesised data.
With respect to input similarity, as shown in Fig.~\ref{fig:sim}, class 3 (Natural Jute Hessian) and category that is incorrectly predicted, such as class 4 (Natural Monkswool) have relatively higher input similarity than other classes which could potentially lead to a mix-up for recognition.

\subsection{Comparison with Other Methods}

\begin{figure}[t]
	\centering
	\includegraphics[width=0.9\columnwidth]{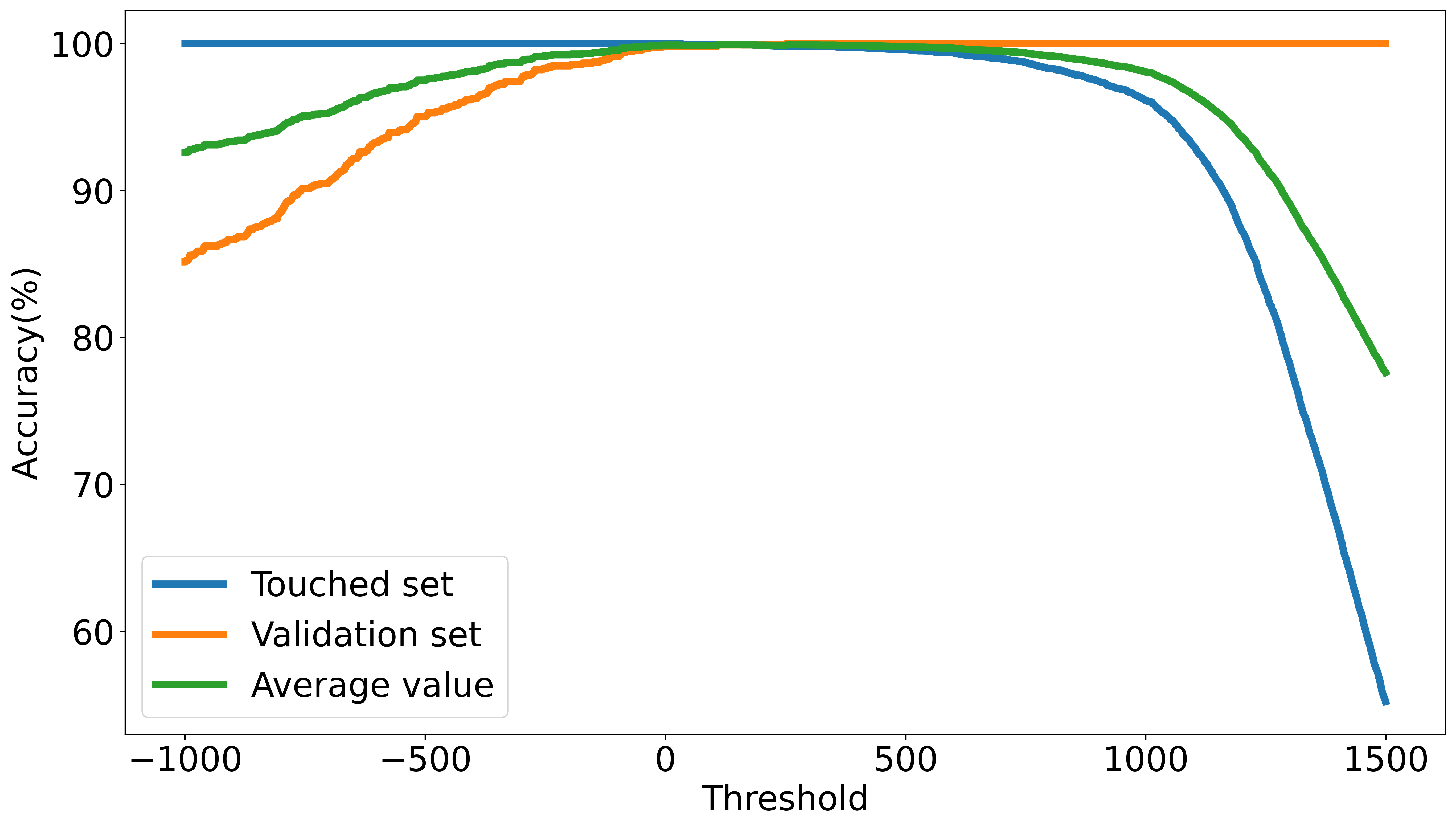}
	\caption{The yellow line represents how much validation data is identified as untouched classes when the threshold is tuned. The blue line represents how much touched data is identified as touched classes when the threshold is tuned. The green line denotes the average value of the two percentages above.   }
	\label{fig:novelty}
	\vspace{-1.2em}
\end{figure}

Here, we first compare our method against other generative model-based methods for the tactile ZSL~\citep{abderrahmane2019deep,lee2019touching}. 
Particularly, we would like to compare our multimodal generative methods with methods based on semantic input~\citep{abderrahmane2019deep} or visual input.
Since there is no prior work using generative models conditioned on visual input, the generative model from~\citep{lee2019touching} is implemented as an alternative, in which the visual images are used to generate the tactile textures. 
In addition to the comparison of generative methods, we also compare our results to those using projection (or mapping) methods, i.e., DAP~\citep{abderrahmane2018haptic} and VT-FC-ZSL~\citep{abderrahmane2018visuo}.
% We test these methods both in ZSL and GZSL settings.
The DAP employs the touched materials to train a projection model to project tactile data to attribute embedding space, then the model is applied on the untouched materials directly for prediction.
The VT-FC-ZSL shares a similar mechanism with DAP, but it uses multimodal input (visual data and tactile data) to train the model and prediction.
To make the model suitable for our tactile textures, a fine-tuned ResNet50 is applied to extract the tactile features from the last pooling layer for the methods in~\citep{abderrahmane2018haptic,abderrahmane2018visuo,abderrahmane2019deep}. 
Note that the methods from~\citep{abderrahmane2018haptic,abderrahmane2018visuo,lee2019touching} do not involve the setup for GZSL. 
We adapt our proposed settings and train a ResNet50 network as $CLS_t$ to simplify the procedure of GZSL.

As detailed in Section~\ref{recognition}, in GZSL, we fit the real tactile features into a Gaussian distribution first. 
Then, if $\log p(x ; \phi_{gau})$ is greater than a selected threshold $\beta$, the input tactile feature $x$ is from touched classes, otherwise, it is from untouched classes. 
Specifically, we use tactile features from $D_{t}$ and $D_{val}$ to tune the threshold $\beta$.
As shown in Fig.~\ref{fig:novelty}, taking the data from the first materials split as an example, we select the threshold $\beta$ that is able to maximise the average accuracy of classifying $x_{t}\in D_{t}$ as touched and $x_{val}\in D_{val}$ as untouched. 
% In~\citep{abderrahmane2019deep}, the semantic attributes are applied to synthesise the tactile features based on the generative model for ZSL problem.
% Specifically, in~\citep{abderrahmane2019deep}, the tactile data are collected by a BioTac sensor, and a 2-layer network was implemented to extract the tactile features. 

% Moreover, we replace the deconvolutional layer and convolutional layer with FC layer in generator and discriminator to achieve a better performance.
%Our tactile data are collected using a GelSight sensor to get high resolution textures (which can be considered as images), so we apply a fine-tuned ResNet50 to extract the tactile features from last pooling layer instead of 2 convolutional layers. Moreover, we replace the deconvolutional layer and convolutional layer with FC layer in generator and discriminator to achieve a better performance.
 
As shown in Table~\ref{tab:compare},
the recognition accuracy of our proposed VS2T-ZSL is $76.36\%$, which is $25.07\%$ and $29.16\%$ higher than the results of other generative methods from~\citep{lee2019touching, abderrahmane2019deep} in ZSL setting, respectively.
A similar trend can be found in the GZSL for both the touched and untouched classes.
Our proposed VS2T-ZSL method has the highest harmonic mean accuracy of $84.27\%$, which shows the superiority of our VS2T-ZSL over other generative methods. 

In the results of projection methods, the use of semantic and visual modalities together in VT-FC-ZSL improves the tactile recognition result largely by $19.64\%$, compared to the result by using semantic attributes only in DAP in ZSL setting. 
In GZSL setting, the harmonic mean of the method VT-FC-ZSL is $15.7\%$ higher than the results of DAP.
However, VS2T-ZSL performs better than both of these projection methods, which demonstrates the effectiveness of our generative model.
% Particularly, using the same multimodal input, the recognition accuracy for ZSL of our proposed method is $13.16\%$ higher than the results using the method VT-FC-ZSL, and the harmonic mean improves by $9.1\%$ in GZSL, 
% Fig.~\ref{fig:shift} (a) (b) show the bias between the projected features and ground-truth semantic embeddings. 
% Different from the similar distributions between the real tactile features and synthesised features with generative method as shown in Fig~\ref{fig:distribution} (a) (b), the projected input features and ground-truth semantic embeddings are kept away, which indicates the effectiveness of our generative method.

\section{Discussion}\label{discussion}
In this section, we discuss several aspects that affect the results of the tactile ZSL. 

\subsection{Unimodal vs. multimodal input for Tactile ZSL}

The application of the multimodal input allows us to measure the objects from different domains, increasing the dimensions to reflect the properties of different objects.
As shown in Fig.~\ref{fig:distribution}, the generated features with multimodal input are closer to the real distribution compared to the generated features with unimodal input, and the boundaries of each class are clearer than the others.
As a result, the prediction of real tactile features will be more accurate using the model trained on the synthesised tactile features with multimodal input.
% In our results, we can observe a synergistic effect with multimodal input and the use of multimodal input achieves a higher recognition results than unimodal input.
% As shown in Fig.~\ref{fig:conf_fused} (a), we demonstrate the confusion matrix with multimodal input. 
% It can be found that the combination of visual and semantic information enables a synergistic effect where the result cannot be achieved with a single modality.
% For example, most tactile features of novel class 3 (\guan{Natural Jute Hessian}, see the 8th column in Fig.~\ref{fig:example}) are misclassified using the semantic input or visual input (as shown in Fig.~\ref{fig:conf_sim} (a) (b)), whereas $35\%$ tactile features are classified correctly with the multimodal input.
% Moreover, as shown in Fig.~\ref{fig:conf_fused} (b), there is still a significant difference in terms of similarity between each class after fusion, compared with the semantic input.

However, due to the implementation of multiple modalities, the computational efficiency decreases compared with the model using single modality. 
% For example, the FLOPS for the generative model of VS2T is xx while the S2T is xx. 
% The recognition accuracy goes up by 20\% at the cost of 20 increase of FLOPS.
For example, $14.36$ millions parameters are trained in the generator for VS2T-ZSL method, while $8.44$ millions parameters are trained for S2T-ZSL method.
% The recognition accuracy for ZSL task goes up by 27.99\% with a increase of 70.14\% parameters in the generator.
Therefore, there is a trade-off between the amount of computation and the accuracy. %A way to speed up time consumption is to apply a more efficient GPU.

\subsection{Visual input vs. semantic input for tactile ZSL}

Compared to multimodal input, unimodal data is much easier to be collected.
Here, we discuss which modality is more effective in tactile ZSL using unimodal input.
While visual images provide an objective measurement that includes fine details of fabrics, semantic attributes offer a high-level description.
Both of them are able to measure the characteristics of target objects from different perspectives. 
However, different from visual images that provide objective measurement with high resolution, the semantic attributes are constrained to a limited number of attributes and are subject to human bias.
From Table~\ref{tab:unimodal}, we can observe that the recognition results using visual input are better than those using semantic input.
It also explains why the visual-based approaches~\citep{abderrahmane2018visuo,lee2019touching} achieve better results than the baseline approaches~\citep{abderrahmane2018haptic,abderrahmane2019deep} that rely only on semantic attributes in Table~\ref{tab:compare}.

Consequently, the visual input is a more objective and accurate auxiliary information to measure the object for tactile ZSL.
A possible way to improve the results of semantic input is to use continuous-valued attributes. 
Compared to binary attributes, continuous-valued attributes can demonstrate the level of properties to improve the recognition results. 
Moreover, a larger number of attributes are expected to increase the dimension of representation.

\subsection{Projection-based vs. generative model-based approaches}
As shown in Table~\ref{tab:compare}, we apply two projection methods~\citep{abderrahmane2018haptic, abderrahmane2018visuo}, two generative model-based methods~\citep{abderrahmane2019deep, lee2019touching}, as well as our proposed multimodal generative method for comparison.
We can see that the performance of the generative methods is better than the projection methods with the same input.
Concretely, the recognition accuracy of our VS2T-ZSL is $13.16\%$ higher than the result from VT-FC-ZSL with multimodal input in ZSL.
The accuracy of~\citep{abderrahmane2019deep} is $3.64\%$ higher than the result from DAP with unimodal input.
It demonstrates the effectiveness of the generative method in tactile ZSL.

For the projection-based method in tactile ZSL, a projection function is learnt between the tactile features and semantic embeddings. 
% \guan{Different from the similar distribution of real tactile features and synthesised features in Fig.~\ref{fig:distribution} (a) (b) by using our proposed generative method, if we learn the projection function only from touched classes and apply it on the untouched classes directly, the embedded features and ground-truth semantic attributes may be kept away due to the bias of the source domain, i.e., domain shift problem~\citep{kodirov2015unsupervised, kodirov2017semantic} (as shown in Fig.~\ref{fig:shift}).}
% For example, the natural union flax (see the 4th column in Fig.~\ref{fig:example}) and herringbone linen (see the 10th column in Fig.~\ref{fig:example}) both have the attribute \textit{rough}, but they have a significant difference in both visual look and tactile texture.
% As a result, it is difficult to predict the \textit{rough} attributes of herringbone linen based on the experience of natural union flax with a  projection model without any adaptation.
However, if the projection function is only learnt from touched classes and is applied to the untouched classes directly, the projected features and semantic embeddings may be kept away due to the bias in touched classes, which is a domain shift problem~\citep{kodirov2015unsupervised, kodirov2017semantic}.
In contrast, the generative model-based method tries to reconstruct the tactile features using the available auxiliary information, e.g., tactile cues embedded in visual images or semantic attributes.
By reconstructing the tactile features, one constraint is met: tactile cues from the visual/semantic domain have to be preserved in the tactile feature generation.
Moreover, due to the fact that the auxiliary information is from the same latent space and is conceptually interlinked, the generator can generate meaningful tactile features for untouched materials, which is able to alleviate the domain shift problem~\citep{ye2021alleviating, wang2018zero}.

\section{Conclusions}\label{conclusions}
In this work, we propose a novel multimodal generative framework to address the tactile ZSL problem. A joint generative model, which integrates the VAE and GAN, is proposed to synthesise the tactile features of untouched materials from visual images and semantic attributes.
In the ablation study, the results demonstrate that our multimodal approach which combines semantic attributes, representing high-level characteristics, and visual images, providing tactile cues from sight, can achieve a synergistic effect, compared to using a single modality.
The extensive experimental results show that the proposed method enables a high recognition accuracy of $83.06\%$ in classifying unknown materials using tactile sensing.
Our proposed VS2T-ZSL method allows the robots to recognise the materials never touched before.

In the future, we will apply the VS2T-ZSL method to different unstructured scenarios where the tactile information is more robust and cannot be easily changed, even if appearance or colour has been altered, e.g., folded clothes. We will also test our methods in other tasks, for example, zero-shot tactile learning for grasping and manipulation of objects with auxiliary information of visual and semantic attributes, to generalise our methods on different tasks.

% With a tactile sensor mounted on the gripper, robots are capable to complete a series of actions: picking up, identifying, and placing, eliminating the need for tactile data collection and training.

Furthermore, in our proposed framework, we use visual images and semantic attributes to generate the data of inaccessible tactile domain. 
Likewise, it is possible to switch the input and output domains, such as using visual and tactile data to synthesise semantic attributes, or using tactile and semantic attributes to synthesise visual data using our framework, which is promising to solve the semantic/visual ZSL problem in a multimodal manner in the future.


{\small
\bibliographystyle{model1-num-names}
\bibliography{cas-refs}
}


\bio{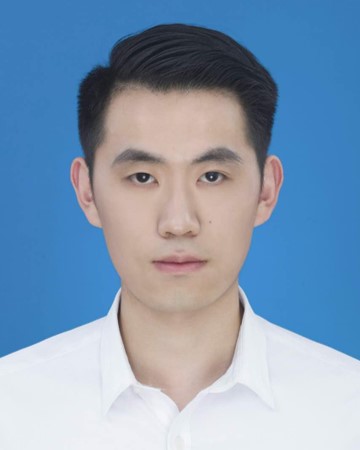}
\textbf{Guanqun Cao} received the B.S. degree from Nanjing Audit University, and the M.Sc. degree from the University of Liverpool. He is currently a Ph.D. candidate in the Department of Computer Science, the University of Liverpool. His research interests include tactile perception and multimodal perception.
\endbio

% \\ \hspace*{\fill} \\
% \\ \hspace*{\fill} \\
\bio{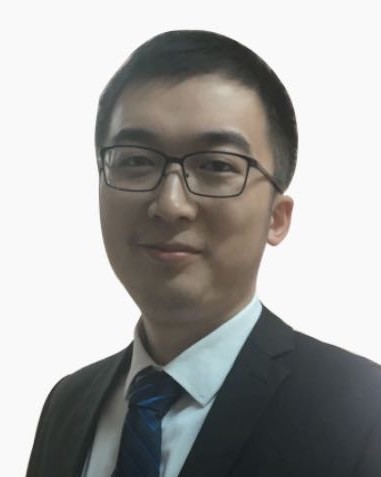}
\textbf{Jiaqi Jiang} received the B.S. and M.S degrees from Beijing Institute of Technology, in 2016 and 2019, respectively.  He is currently a Ph.D. candidate in the Department of Engineering, King's College London. He was a Ph.D. candidate in the Department of Computer Science, the University of Liverpool. His research interests include robot grasping and sensory synergy of vision and touch.
\endbio
% \\ \hspace*{\fill} \\
% \\ \hspace*{\fill} \\
% \\ \hspace*{\fill} \\
% \\ \hspace*{\fill} \\
\bio{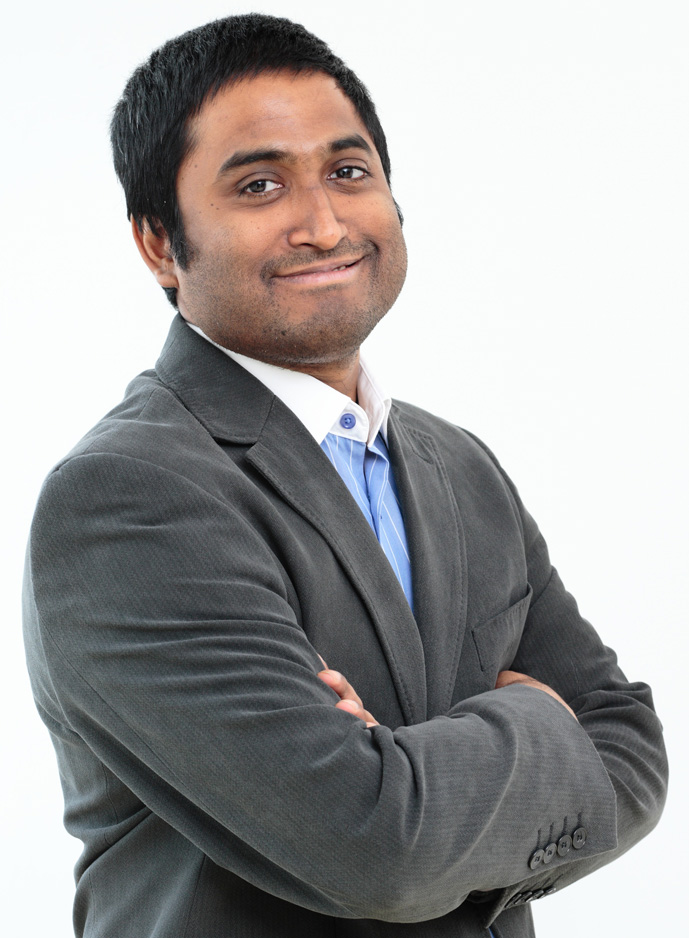}
\textbf{Danushka Bollegala} is a Professor at the Department of Computer Science, University of Liverpool. He is also an Amazon Scholar. He obtained his PhD in 2009 from the University of Tokyo, Japan. He has received many prestegious awards such as the IEEE Young Author Award, GECCO best paper award and JSAI best journal paper award. He has published over 170 peer-reviewed papers in top international venues in NLP, ML and AI.
\endbio

\bio{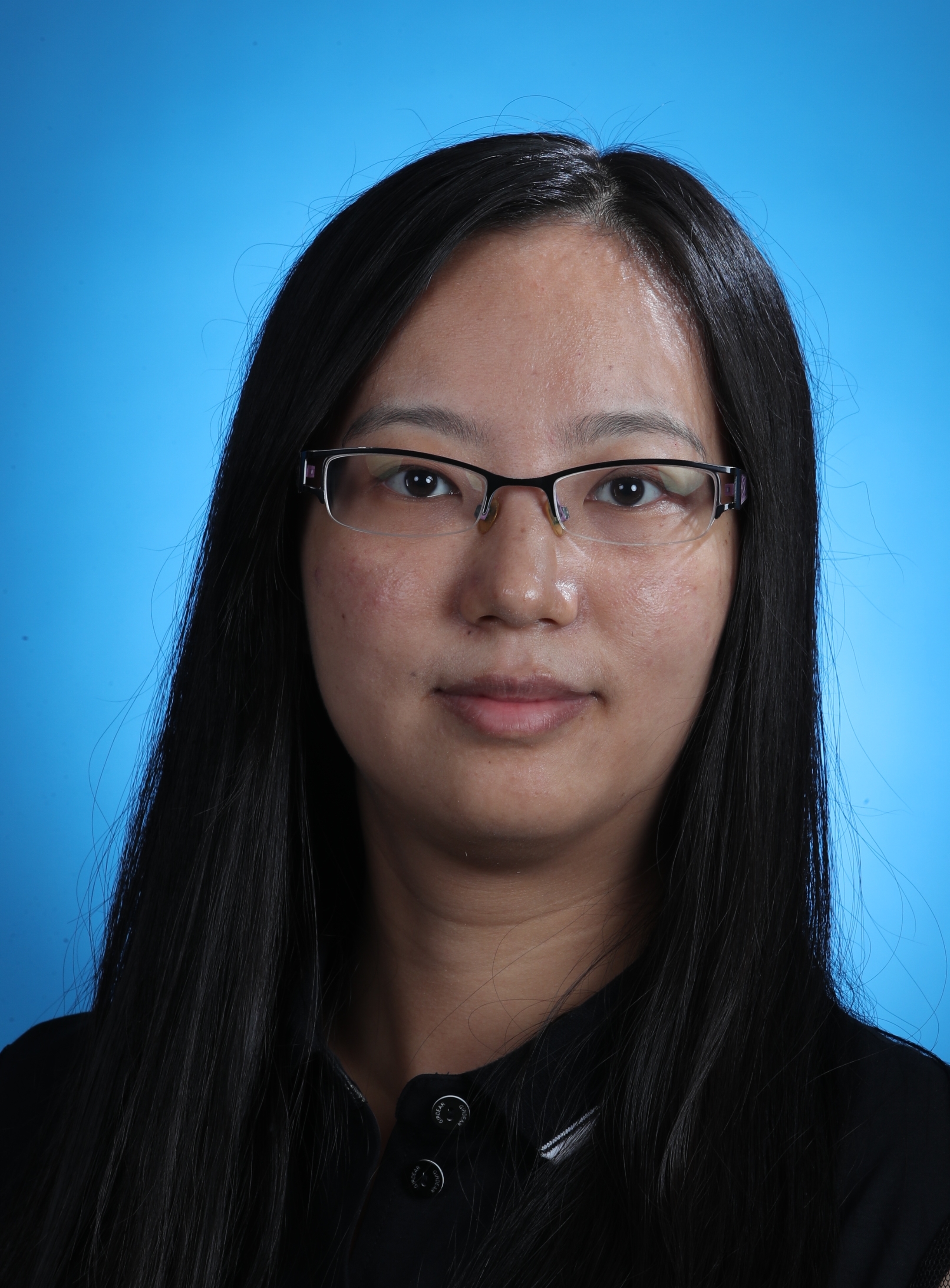}
\textbf{Min Li} is an Associate Professor at the  School of Mechanical Engineering, Xi'an Jiaotong University, China. She received her B.Sc. degree in Mechanical Engineering and her M.Sc. degree in Agricultural Mechanization Engineering from Northwest A$\&$F University, China, in 2007 and 2010, respectively. She was awarded the Ph.D. degree in Robotics at King's College London, UK, in 2014. From 2015 to 2017, she was a Lecturer with Xi'an Jiaotong University. Her research interests include haptic feedback for robots, soft robots, rehabilitation robots.
\endbio

\bio{bio/ShanLuo.jpg}
\textbf{Shan Luo} is a Senior Lecturer (Associate Professor) at the Department of Engineering, King’s College London. Previously, he was a Lecturer at the University of Liverpool, and Research Fellow at Harvard University and University of Leeds. He was also a Visiting Scientist at the Computer Science and Artificial Intelligence Laboratory (CSAIL), MIT. He received the B.Eng. degree in Automatic Control from China University of Petroleum, Qingdao, China, in 2012. He was awarded the Ph.D. degree in Robotics from King’s College London, UK, in 2016. His research interests include tactile sensing, robot learning and robot visual-tactile perception.
\endbio

\end{document}